%% file: modelcapacity.tex
\pdfoutput=1
\documentclass[11pt]{article}  %

\usepackage[small]{caption}
\usepackage{subcaption}
\usepackage{amsfonts}       %
\usepackage[final]{naacl2021}
\usepackage{svg}
\usepackage{graphicx} %
\usepackage{natbib}
\usepackage{appendix}

\usepackage{amssymb}%
\usepackage{pifont}%
\newcommand{\cmark}{\ding{51}}%
\newcommand{\xmark}{\ding{55}}%

\usepackage[utf8]{inputenc} %
\usepackage[T1]{fontenc}    %
\usepackage{url}            %
\usepackage{booktabs}       %
\usepackage[shortlabels]{enumitem}
\setlist{nosep}
\usepackage{nicefrac}       %
\usepackage[activate=true,
    final,
    tracking=false,   %
    kerning=true,
    spacing=true,
    factor=1050,
    stretch=30,
    shrink=30]{microtype}      %
\usepackage{xpatch}
\xapptocmd\normalsize{%
 \abovedisplayskip=11pt plus 3pt minus 9pt
 \abovedisplayshortskip=0pt plus 3pt
 \belowdisplayskip=11pt plus 3pt minus 9pt
 \belowdisplayshortskip=6.5pt plus 3.5pt minus 3pt
}{}{}
\usepackage[]{comment}
\usepackage[ruled,vlined]{algorithm2e}

\usepackage{amsmath}
\usepackage{amssymb}
\usepackage{mathtools}
\usepackage[thinc]{esdiff}
\usepackage{xspace}
\usepackage{newtxtext}
\usepackage{tensor}
\usepackage{amsthm}
\usepackage{multirow}
\usepackage[smallerops]{newtxmath}

\usepackage{bm}
\usepackage{thmtools}
\usepackage[noabbrev,capitalize]{cleveref} %

\crefname{equation}{equation}{equations}   %
\crefname{footnote}{footnote}{footnotes}   %
\crefname{section}{\S}{\S\S}
\Crefname{section}{\S}{\S\S}    %
\crefformat{section}{#2\S#1#3}  %
\Crefformat{section}{#2\S#1#3}
\crefrangeformat{section}{\S\S#3#1#4--#5#2#6}
\Crefrangeformat{section}{\S\S#3#1#4--#5#2#6}
\crefformat{section}{#2\S#1#3}  %
\crefrangeformat{section}{\S\S#3#1#4--#5#2#6}

\usepackage{tikz}
\usepackage{pgfplots}
\pgfplotsset{compat=1.14}
\usepackage{floatrow}

\usepackage[disable]{todonotes}
\makeatletter
\newcommand*\iftodonotes{\if@todonotes@disabled\expandafter\@secondoftwo\else\expandafter\@firstoftwo\fi}  %
\makeatother

\newcommand{\mycomment}[1]{%
}%

\usepackage{marginnote}

\newlength{\extramargin}
\usepackage{geometry}
\usepackage{calc}
\iftodonotes{\newgeometry{paperwidth=\paperwidth+\extramargin*2,marginparwidth=\marginparwidth+\extramargin,width=\textwidth}}{}

\newcommand{\cutforspace}[1]{%
}%

\usepackage{bm}
\newcommand{\vecc}[1]{{\boldsymbol{\mathbf{#1}}}}

\newcommand{\vx}{{\vecc{x}}}
\newcommand{\vy}{{\vecc{y}}}

\newcommand{\va}{{\vecc{a}}}
\newcommand{\vb}{{\vecc{b}}}

\newcommand{\vd}{{\vecc{d}}}

\newcommand{\vh}{{\vecc{h}}}

\newcommand{\vz}{{\vecc{z}}}

\newcommand{\vtheta}{\vecc{\theta}}
\newcommand{\vTheta}{\vecc{\Theta}}
\newcommand{\vphi}{\vecc{\phi}}
\newcommand{\Z}{Z_{\vtheta}}

\newcommand{\g}{g_{\vtheta}}

\newcommand{\eos}{\texttt{\$}\xspace}
\newcommand{\sep}{\texttt{\#}\xspace}

\DeclareMathOperator*{\argmax}{argmax}
\usepackage{stmaryrd}  %

\usepackage[safe]{tipa}

\newcommand{\defn}[1]{\textbf{#1}}   %
\newcommand{\Real}{\mathbb{R}}
\newcommand{\Bool}{\mathbb{B}}
\newcommand{\E}[2][]{\mathop{\mathbb{E}}_{{#1}}[#2]}   %
\newcommand{\hfrac}[2]{#1/#2}

\newcommand{\Xc}{L}

\newcommand{\Zc}{S}
\newcommand{\vxhat}{\hat{\vx}}
\newcommand{\vzhat}{\hat{\vz}}

\newcommand{\tildep}{\tilde{p}}
\newcommand{\ptilde}{\tilde{p}}
\newcommand{\qtilde}{\tilde{q}}
\newcommand{\rtilde}{\tilde{r}}
\newcommand{\prefixof}{\preceq}

\newcommand{\Mq}{M^{\mathrm{q}}}
\newcommand{\Mr}{M^{\mathrm{r}}}
\newcommand{\Mtildep}{M^{\mathrm{\tilde{p}}}}

\newcommand{\Mtilder}{M^{\mathrm{\tilde{r}}}}
\newcommand{\vThetap}{\vTheta^{\mathrm{p}}}
\newcommand{\vThetaq}{\vTheta^{\mathrm{q}}}

\newcommand{\vThetatr}{\vtheta^{\mathrm{\tilde{r}}}}
\newcommand{\vthetapn}{\vtheta^{\mathrm{p}}_n}
\newcommand{\vthetaqn}{\vtheta^{\mathrm{q}}_n}

\newcommand{\vthetaq}{\vtheta^{\mathrm{q}}}

\newcommand{\vthetatr}{\vtheta^{\mathrm{\tilde{r}}}}
\newcommand{\npppoly}{{\mathrm{NP} \nsubseteq \mathrm{P/poly}}}
\usepackage{url}
\newcommand{\defeq}{\triangleq}
\newcommand{\approptoinn}[2]{\mathrel{\vcenter{
  \offinterlineskip\halign{\hfil$##$\cr
    #1\propto\cr\noalign{\kern2pt}#1\sim\cr\noalign{\kern-2pt}}}}}

\crefname{ineq}{inequality}{inequalities}
\creflabelformat{ineq}{#2{\upshape(#1)}#3} 
\newtheorem{theorem}{Theorem}

\crefname{localtheorem}{Theorem}{Theorems}
\numberwithin{localtheorem}{section}

\crefname{locallemma}{Lemma}{Lemmas}
\numberwithin{locallemma}{section}

\usepackage{thm-restate}
\usepackage{siunitx}
\usepackage{ifthen}
\newboolean{shortver}
\setboolean{shortver}{true}%
\DeclareRobustCommand{\thinskip}{\hskip 0.16667em\relax}
\def\emdash{---}
\def\d@sh#1#2{\unskip#1\thinskip#2\thinskip\ignorespaces}
\def\Dash{\d@sh\nobreak\emdash}
\def\Ldash{\d@sh\empty{\hbox{\emdash}\nobreak}}
\def\Rdash{\d@sh\nobreak\emdash}

\title{Limitations of Autoregressive Models and Their Alternatives}

\author  
{
	\begin{tabular}{ccccc}
		Chu-Cheng Lin\raise1.0ex\hbox{\normalfont\normalsize $\sharp$}\Thanks{Part of this work was done at Facebook AI.} & Aaron Jaech\raise1.0ex\hbox{\normalfont\normalsize $\flat$} & Xin Li\raise1.0ex\hbox{\normalfont\normalsize $\sharp$} & Matthew R. Gormley\raise1.0ex\hbox{\normalfont\normalsize $\natural$} & Jason Eisner\raise1.0ex\hbox{\normalfont\normalsize $\sharp$}
	\end{tabular}
	\\
	\raise1.0ex\hbox{\normalsize $\sharp$}Department of Computer Science, Johns Hopkins University\\
	\raise1.0ex\hbox{\normalsize $\flat$}Facebook AI\\
	\raise1.0ex\hbox{\normalsize $\natural$}Machine Learning Department, Carnegie Mellon University\\
	{\tt \{kitsing,lixints,jason\}@cs.jhu.edu ajaech@fb.com mgormley@cs.cmu.edu}
}

\begin{document}

\maketitle
\begin{abstract}
	Standard autoregressive language models perform only polynomial-time computation to compute the probability of the next symbol.  While this is attractive, it means they cannot model distributions whose next-symbol probability is \emph{hard} to compute. Indeed, they cannot even model them well enough to solve associated \emph{easy} decision problems for which an engineer might want to consult a language model.%
	  These limitations apply no matter how much computation and data are used to train the model, unless the model is given access to oracle parameters that grow \emph{superpolynomially} in sequence length.

	  Thus, simply training larger autoregressive language models is not a panacea for NLP.  
	Alternatives include energy-based models (which give up efficient sampling) and latent-variable autoregressive models (which give up efficient scoring of a given string).  Both are powerful enough to escape the above limitations.%
	 
\end{abstract}
\input{newintro}

\input{background}

\input{s3}

\input{sidestepping}

\input{related}

\input{conclusion}
\section*{Acknowledgements}

We thank the anonymous reviewers for their comments.
We also thank our colleagues at Johns Hopkins University, Facebook, and Carnegie Mellon University for their comments on earlier versions of the manuscript.
This material is based upon work at Johns Hopkins University supported by the National Science Foundation under Grant No.\@ 1718846. It does not represent the views of Microsoft (where Dr.\@ Eisner is also a paid employee, in an arrangement that has been reviewed and approved by the Johns Hopkins University in accordance with its conflict of interest policies).

\bibliographystyle{acl_natbib}
\bibliography{nips}
\clearpage
\appendix
\input{appendix}
\end{document}

%% file: newintro.tex
\section{Introduction}
\label{sec:intro}

\begin{table*}[ht]
	\centering
	\resizebox{\textwidth}{!}{%
		\begin{tabular}{m{.56\textwidth}cccl}
			\toprule
			\multicolumn{1}{c}{Model family} &
			\multicolumn{1}{m{.13\textwidth}}{\ifthenelse{\boolean{shortver}}{\small}{} Compact \mbox{parameters}?} &
			\multicolumn{1}{m{.12\textwidth}}{\ifthenelse{\boolean{shortver}}{\small}{} Efficient \mbox{scoring}?} &
			\multicolumn{1}{m{.15\textwidth}}{\ifthenelse{\boolean{shortver}}{\small}{} Efficient sampling and normalization?} &
			\multicolumn{1}{c}{Support can be \ldots } \\ 
			\cmidrule(lr){1-1} \cmidrule(lr){2-4} \cmidrule(lr){5-5}
			\ifthenelse{\boolean{shortver}}{\small}{} ELN/ELNCP: Autoregressive models (\cref{sec:local-normalization})       & \cmark & \cmark & \cmark & \emph{some but not all} $L \in \mathrm{P}$ \\
			\ifthenelse{\boolean{shortver}}{\small}{} EC/ECCP: Energy-based models   (\cref{sec:ebm})                & \cmark & \cmark & \xmark & \emph{all} $L \in \mathrm{P}$ but \emph{no} $L \in \mathrm{NPC}$ \\
			\ifthenelse{\boolean{shortver}}{\small}{} Lightly marginalized ELNCP: { Latent-variable autoregressive models} (\cref{sec:marginalized}) & \cmark & \xmark & \cmark & \emph{all} $L \in \mathrm{NP}$ \\ %
						\ifthenelse{\boolean{shortver}}{\small}{} Lookup models (\cref{sec:semiparametric-models})          & \xmark & \cmark & \cmark & \emph{anything} \\ 
						\bottomrule
		\end{tabular}%
	}
	\caption{\small A feature matrix of parametric model families discussed in this paper. Also see \cref{fig:modelsupportzoo} in the appendices.}
	\label{fig:modelfamilycomparison}
\end{table*}\ifthenelse{\boolean{shortver}}{\begin{figure}[ht]}{\begin{figure*}[ht]}
	\centering
	\ifthenelse{\boolean{shortver}}{\includegraphics[width=.964\linewidth]{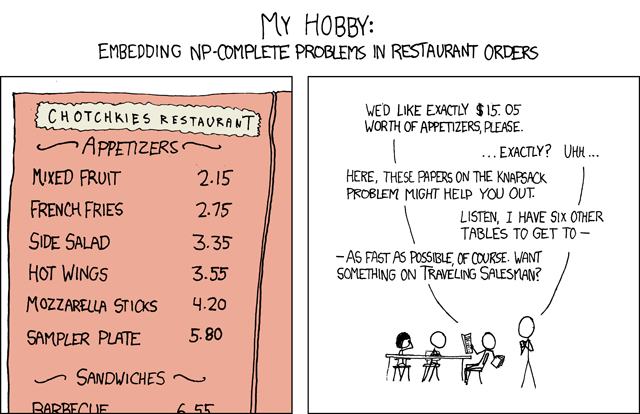}}{	\includegraphics[width=1.\linewidth]{figures/np_complete.png}}
	\caption{\ifthenelse{\boolean{shortver}}{\small}{}
		Valid answers to hard natural language inference problems %
		can be hard to find \cite{xkcd}, but in many cases can be checked efficiently (\emph{e.g.} the \textsc{Knapsack} problem in the comic). Given a \emph{large enough} parametric autoregressive model with \emph{correct} parameters, we can efficiently solve \emph{all} problem instances
		with input length $n$, and efficiently verify the solutions \Dash but the required model size can grow superpolynomially in $n$.
		(This allows the model to store precomputed results that we can look up in $O(n)$ at test time.)
		A main observation of this paper is that assuming $\npppoly$,
			 then without such a superpolynomial growth in model size, autoregressive models cannot even be used to verify 
			 answers to some problems where polynomial-time verification
			 algorithms do exist.
	}
	\label{fig:xkcd}
\ifthenelse{\boolean{shortver}}{\end{figure}}{\end{figure*}}
Sequence modeling is a core NLP problem. Many sequence models $\ptilde$ are efficient at \emph{scoring strings}: given a string $\vx$, its score $\ptilde(\vx)$ can be computed in $O(\mathrm{poly}(|\vx|))$.  For example, an RNN \cite{mikolov2011rnnlm}
scores $\vx$ in time $O(|\vx|)$ while a Transformer \cite{Vaswani2017AttentionIA} does so in time $O(|\vx|^2)$.
The score may be an unnormalized probability, and can be used to rank candidate strings.

\iftodonotes{\clearpage}

Many sequence models also make it easy to compute marginal properties
of $\ptilde$.  They support efficient \emph{sampling} of
strings $\vx$ (which allows unbiased approximation of marginal
expectations).  And they support efficient computation of the \emph{normalizing
  constant} $Z = \sum_\vx \ptilde(\vx)$ (or simply guarantee $Z=1$) for any value of the model
parameters.\ifthenelse{\boolean{shortver}}{\looseness=-1}{%
}

How about training?  Briefly: If a sequence model can efficiently compute
$\ptilde(\vx)$ (and its derivatives with respect to model parameters),
then it is efficient to compute parameter updates for
noise-contrastive estimation \cite{Gutmann2010NoisecontrastiveEA,gutmann-12-nce} or score-matching \cite{JMLR:v6:hyvarinen05a}.  If sampling $\vx$ or
computing $Z$ (and its derivatives) is \emph{also} efficient, then it is efficient to
compute parameter updates for ordinary MLE training.

Finally, popular sequence models are \emph{compact}.  Usually a fixed-size model is used to score strings $\vx$ of all lengths.  More generally, it might be reasonable to use an $O(\mathrm{poly}(n))$-sized parameter vector $\vtheta_n$ when $\vx$ has length $n$, at least if parameter vectors can be obtained (perhaps from an oracle) for all needed lengths.  In this paper, we investigate what can and cannot be achieved with models that are compact in this sense.  This setup allows us to discuss the asymptotic behavior of model families.

\defn{Standard autoregressive models} %
have the form $p(\vx) = \prod_t p(x_t \mid \vx_{<t})$\footnote{In this paper we use the shorthand $\vx_{<t} \triangleq x_1 \ldots x_{t-1}$.} where each factor is efficient to compute
from a fixed parameter vector.  These models satisfy all three of the desiderata above.
By using flexible neural network architectures, standard autoregressive models have achieved stellar empirical results in many applications  \citep{oord2016wavenet,child2019generating,zellers2019defending,brown2020language}. However there are still tasks that they have not mastered: \emph{e.g.}, it is reported that they struggle at deep %
logical structure, even when initialized to huge pretrained models \cite{wang2019glue}.

We point out that, unfortunately, there are certain sequence distributions whose unnormalized string probabilities $\ptilde(\vx)$ are \emph{easy} to compute individually,
yet 
\ifthenelse{\boolean{shortver}}{whose autoregressive factors $p(x_t \mid \vx_{<t})$ are \emph{$\mathrm{NP}$-hard} 
    to compute or even approximate, or are even \emph{uncomputable}.
	Thus, standard autoregressive models are \emph{misspecified} for these distributions (cannot fit them). It does not help much to focus on strings of bounded length, or to enlarge the model: under the common complexity-theoretic assumption ${\npppoly}$, the parameter size $|\vtheta_n|$ must grow \emph{superpolynomially} in $n$ to efficiently approximate the probabilities of all strings of length up to $n$. 
}{the autoregressive factors $p(x_t \mid \vx_{<t})$ are \emph{hard} to compute, or even \emph{uncomputable}. In other words, there is no standard autoregressive model that can exactly match the support of $\ptilde$, let alone parametrize it. While (un)computability may not concern many machine learning applications where we only need to fit finite datasets, and/or can afford to approximate more complicated distributions with larger models, we further show that 
there are sequence distributions that \emph{no} autoregressive model will scale up well to approximate:
their unnormalized string probabilities $\ptilde(\vx)$ are again easy to compute, but that computing the autoregressive factors is $\mathrm{NP}$-hard. We will show that under the common complexity-theoretic assumption $\npppoly$, the model parameter size $|{\vtheta}_n|$ must grow \emph{superpolynomially} in $n$ to approximate the probabilities of all strings of length up to $n$. In other words, standard autoregressive models are \emph{misspecified} for these distributions.
}

Indeed, one of our main findings is that there exist unweighted languages $L \in \mathrm{P}$
for which \emph{no} standard autoregressive model has $L$ as its support, \emph{i.e.}, assigns weight $> 0$ to just the strings $\vx \in L$.
This is downright depressing, considering the costs invested in training huge parametric autoregressive models \cite{parrots2021}.  Since $L\in \mathrm{P}$, it is trivial to build an efficient scoring function $\ptilde(\vx)$ with fixed parameters that has $L$ as its support \Dash just not an autoregressive one. 
The problem holds %
	for \emph{all} standard autoregressive models, 
	regardless of how much computation and training data are used to learn the model parameters.

That is, for an $\mathrm{NP}$-hard problem, scoring a string $\vx$ under a standard autoregressive model $p(\vx)$ cannot be used to \emph{verify} a witness.
Nor can \emph{finding} a witness be solved by prompting such a model with a description of a problem instance and sampling a continuation $\vx$ of that string.
Such problems are abundant in NLP: for example, surface realization under Optimality Theory \cite{Idsardi2006ASP}, decoding text from an AMR parse \cite{cai-knight-2013-smatch}, phrase alignment between two sentences \cite{DeNero2008TheCO}, and in general inference for propositional logic \cite{cook1971}, which underlies the $\mathrm{NP}$-hardness of general natural language inference, as in \cref{fig:xkcd}. In other words, our results imply that standard autoregressive models do not have the right structure to capture important linguistic regularities: \emph{e.g.}, that observed sequences were \emph{in fact} constructed to be phonologically optimal, expressive of a semantic form, or logically coherent!

Our work is also relevant to autoregressive models of fixed-dimensional vectors, such as NADE \cite{uria2016neural}.  These can be extended to arbitrary $n$-dimensional vectors by providing separate parameters $\vtheta_n$ for each $n$.  Our constructions imply that for some distributions, $|\vtheta_n|$ must grow superpolynomially in $n$, even though this would be not be necessary if the models were not autoregressive.

In the remainder of this paper, we formalize our three desiderata for sequence models.  We formalize compact autoregressive models and describe some limitations on their expressiveness.
We then show that it can help to choose an alternative model family that relaxes any one of the three desiderata (\cref{fig:modelfamilycomparison}).

%% file: background.tex
\section{Background}
\label{sec:background}
\subsection{Weighted languages}
\label{sec:weighted-languages}
An \defn{unweighted language} $\Xc \subseteq V^*$ is a set of strings $\vx$ over a finite alphabet $V$. A \defn{weighted language} $\tilde{p}$ is a function $\tilde{p}: V^* \rightarrow \Real_{\geq 0}$.  It may be regarded as specifying an unweighted language $\Xc = \mathrm{support}(\tilde{p}) \defeq \{\vx: \tilde{p}(\vx) \neq 0\}$ along with positive weights for the strings in $\Xc$. 
We say that a weighted language $\tilde{p}$ is \defn{normalizable} if its \defn{global normalizing constant} $Z \triangleq \sum_{\vx \in V^*} \tilde{p}(\vx)$ is finite and strictly positive. When $\tilde{p}$ is normalizable, $p(\vx) \triangleq \hfrac{\tilde{p}( \vx )}{ Z }$ is a probability distribution over $\Xc$.  A \defn{distribution} is any weighted language whose global normalizing constant is 1.\ifthenelse{\boolean{shortver}}{\looseness=-1}{%
}

Let $\hat{\vx} \prefixof \vx$ mean
that $\hat{\vx}$ is a prefix of $\vx \in V^*$ (not necessarily a strict prefix).  If $\tilde{p}$ is normalizable, then 
$Z(\hat{\vx}) \triangleq  \sum_{\vx \in V^*: \hat{\vx} \prefixof \vx} \tilde{p}(\vx)$ is $\leq Z$ for any $\hat{\vx} \in V^*$, yielding a marginal \defn{prefix probability} $\hfrac{Z(\hat{\vx})}{Z}$.  
If the prefix $\hat{\vx}$ has positive prefix probability, then it admits a \defn{local conditional probability} $p(x \mid \hat{\vx}) \triangleq \hfrac{Z(\hat{\vx}\,x)}{Z(\hat{\vx})}$ for each symbol $x \in V$, where the denominator is interpreted as a \defn{local normalizing constant}.  This is the conditional probability that if a random string starts with the prefix $\hat{\vx}$, the next symbol is $x$.  There is also a probability $p(\eos \mid \hat{\vx}) \triangleq 1 - \sum_{x\in V} p(x \mid \hat{\vx}) = \tilde{p}(\hat{\vx})/Z(\hat{\vx}) \geq 0$ that the string ends immediately after $\hat{\vx}$; the special symbol $\eos \notin V$ represents ``end of string.''

\subsection{Computation for weighted languages}
\label{sec:computation-for-weighted-languages}
We define a weighted language $\tilde{p}$ to be \defn{computable} if it is defined by a Turing machine (also called $\tilde{p}$) that maps any $\vx \in V^*$ to $\tilde{p}(\vx) \in \mathbb{Q}_{\geq 0}$ in finite time. The Turing machine does not have to compute $Z$.

While the computable weighted languages allow any computable function as $\tilde{p}$, most architectures for defining weighted languages (\emph{e.g.}, RNNs or Transformers) do only a bounded or linear 
amount of work per input symbol.  
As a result, they compute $\tilde{p}(\vx)$ in time $O(\mathrm{poly}(|\vx|))$ (that is, $\tilde{p} \in \mathrm{FP}$).
We refer to such weighted languages as \defn{efficiently computable} (\defn{EC}). 
This does not imply that the normalized version $p$ is efficiently computable,
since finding the denominator $Z$ requires summing over all of $V^*$.

If we tried to construct the same normalized distribution $p$ as in the previous paragraph using a standard autoregressive model, we would model it as a product of local conditional probabilities, $p(\vx) = (\prod_{t=1}^{|\vx|} p(x_t \mid \vx_{<t}) ) p(\eos \mid \vx)$. Most such architectures again do only a bounded or linear amount of work per input symbol.  Yet one suspects that this may not always be enough work to do the job: the local conditional probabilities of the original $\tilde{p}$ are expensive to compute (unless $\tilde{p}$ has some special structure making $Z(\vxhat)$ tractable). 
Indeed, the observation of this paper is that for some efficiently computable weighted languages $\tilde{p}$, the local conditional probabilities are expensive to compute or even to approximate well. More precisely, autoregressive models cannot fit the local conditional probabilities unless they are superpolynomial in either their runtime or in their number of parameters (where the parameters may be precomputed at training time).
We now explain how to formalize these notions.

\subsection[complexityclasses]{Non-uniform computation}
\label{sec:compact} %

In the machine learning approach to sequence modeling, we usually do not manually design the Turing machine behind $\tilde{p}$. Rather, we design a model $M$ with \emph{parameters} $\vtheta$.  $M$ is a Turing machine that reads $\vtheta$
and outputs a specialized Turing machine 
$\tilde{p}_{\vtheta} \triangleq M(\vtheta)$ that can score strings $\vx$ and hence defines a weighted language.
Without loss of generality, we will express $\vtheta$ as a string in $\mathbb{B}^*$ (where $\mathbb{B} \defeq \{0,1\}$).
For each $\vtheta$, we obtain a potentially different weighted language.%

Strings vary in length, and accurate modeling of longer strings may sometimes require more complex computations with more parameters.  For example, when $V$ is a natural language alphabet, a recurrent neural network may require more hidden units to model sentences of the language rather than individual words, and even more units to model whole documents.  To accommodate this, we allow an \emph{infinite sequence} of parameter vectors, $\vTheta = \{ \vtheta_n \in \mathbb{B}^* \mid n \in \mathbb{N} \}$, which yields an infinite sequence of Turing machines $\{ \tilde{p}_n \mid n \in \mathbb{N}\}$ via $\tilde{p}_n \triangleq M(\vtheta_n)$.
We then define $\tilde{p}_{\vTheta}(\vx) \triangleq \tilde{p}_{|\vx|}(\vx)$, so a string of length $n$ is scored by the $\tilde{p}_n$ machine.  This is known as \defn{non-uniform computation}.
Of course, it is legal (and common) for all of the $\vtheta_n$ to be equal, or empty, but if desired, we can obtain more power by allowing the number of parameters to grow with $n$ if needed.  

We can now consider \emph{how rapidly} the parametric and runtime complexity may grow. 
\begin{itemize}[leftmargin=*]
\item If $|\vtheta_n|$ is permitted to grow exponentially, then one can fit \emph{any} weighted language $\tilde{p}$ (even an uncomputable one).\footnote{See our remark on computability  in \cref{sec:proofs}.}  Simply use $\vtheta_n$ to encode a trie with $O(|V|^{n+1})$ nodes that maps $\vx \mapsto \tilde{p}(\vx)$ for any $|\vx|$ of length $n$, and design $M$ such that the Turing machine $\tilde{p}_n = M(\vtheta_n)$ has a (large) state transition table that mirrors the structure of this trie. The resulting collection of Turing machines $\{ \tilde{p}_n \mid n \in \mathbb{N}\}$
can then compute $\tilde{p}(\vx)$ exactly for any $\vx$,  with only linear runtime $O(|\vx|)$ (which is used to traverse the trie). 
\item Separately, if unbounded runtime is permitted for $M$, then one can exactly fit \emph{any computable} weighted language $\tilde{p}$. Simply have $M$, when run on $\vtheta_n$, \emph{compute} and return the large trie-structured 
$\tildep_n$ that was mentioned above.  In this case, $M$ need not even use the parameters $\vtheta_n$, except to determine $n$. 
\item Finally, if unbounded runtime is permitted for $\ptilde_n$, then again one can exactly fit \emph{any computable} weighted language $\ptilde$.  In this case, $M$ trivially returns $\ptilde_n = \ptilde$ for all $n$.
\item However, if the parameters $\vTheta$ are ``compact'' in the sense that $|\vtheta_n|$ grows only as $O(\mathrm{poly}(n))$, and also $\ptilde_n = M(\vtheta_n)$ is constructed by $M$ in time $O(\mathrm{poly}(n))$, and $\ptilde_n$ scores any $\vx$ of length $n$ in time $O(\mathrm{poly}(n))$, then we say that the resulting weighted language $\tilde{p}$
is \defn{efficiently computable with compact parameters} (ECCP).\footnote{Since we require $M$ to run in polytime, it can only look at a polynomial-sized portion of $\vtheta_n$.  Hence it is not really crucial for the parameters $\vthetapn$ to be compact, but we nonetheless include this intuitive condition, without loss of generality.}
  We refer to $M$ paired with a parameter space of possible
  compact values for $\vTheta$ as an \defn{ECCP model}.
\end{itemize}

Neural models of weighted languages are typically ECCP models.%
  The construction and execution of the neural network $\ptilde_n$ may perform a polynomial amount of total computation to score the string $\vx$.  This computation may involve parameters that were precomputed using any amount of effort (\emph{e.g.}, training on data) or even obtained from an oracle (they need not be computable).  However, the exponentially many strings of length $n$ must share a polynomial-size parameter vector $\vtheta_n$, which prevents the solution given in the first bullet point above.

  In practice one takes $\vtheta_n = \vtheta$ for all $n$ and obtains $\vtheta \in \mathbb{R}^d$ by training.
  However, we do not consider whether such parameters are easy to estimate or even computable.  We simply ask, for a given target language $\ptilde$, whether there \emph{exists} a polynomially growing sequence $\vTheta$ of ``good'' parameter vectors for any parametric model $M$.  When not, there can be no scheme for estimating arbitrarily long finite prefixes of such a sequence.  So for any polynomial $f$, any training scheme that purports to return a trained model of size $f(n)$ that works ``well'' for strings of length $\leq n$ must fail for large enough $n$\Dash even if unlimited data, computation, and 
  oracles are allowed at training time.

\subsection{$\mathrm{P}$, $\mathrm{P/poly}$, and $\mathrm{NP/poly}$}
\label{sec:p-poly}

The phrase ``efficiently computable with compact parameters'' means that without access to those parameters, the ECCP weighted language may no longer be efficiently computable.  Indeed, it need not be computable at all, if the parameter vectors store the outputs of some uncomputable function.

Our definitions above of EC and ECCP weighted languages are weighted generalizations of complexity classes $\mathrm{P}$ and $\mathrm{P/poly}$, respectively,\footnote{%
	Namely the nonnegative functions in $\mathrm{FP}$ and $\mathrm{FP/poly}$.} 
and their supports are always unweighted languages in $\mathrm{P}$ and $\mathrm{P/poly}$, respectively.
An unweighted language $L$ is in $\mathrm{P}$ iff there is a deterministic Turing machine that decides in $O(\mathrm{poly}(|\vx|))$ time  whether $\vx \in L$. And an unweighted language $L'$ is in $\mathrm{P/poly}$ iff%
\footnote{Our presentation of $\mathrm{P/poly}$ is a variant of \citet[\S 6]{arora2009computational}, in which inputs $\vx$ of length $n$ are evaluated by a polytime function $M$ that is given an advice string $\vtheta_n$ as an auxiliary argument.  This corresponds to a neural architecture $M$ that can consult trained parameters $\vtheta_n$ at runtime.  We have replaced the standard call $M(\vtheta_n,\vx)$ with the ``curried'' expression $M(\vtheta_n)(\vx)$, which we still require to execute in polynomial total time.  Here the intermediate result $M_n = M(\vtheta_n)$ corresponds to a trained runtime model for inputs of length $n$.  Our Turing machines $M_n$ have size polynomial in $n$ (because they are constructed by $M$ in polynomial time).  They correspond to the polynomial-sized boolean circuits $M_n$ that are used to evaluate inputs of length $n$ under the classical definition of $\mathrm{P/poly}$ \cite{ladner1975}.  We exposed these intermediate results $M_n$ only to observe in \cref{sec:compact} and \cref{sec:semiparametric-models} that if we had allowed the $M_n$ to grow exponentially, they would have been able to encode the answers in tries.}
there exist Turing machines $\{M_n: n \in \mathbb{N}\}$ such that $M_n$ decides in $O(\mathrm{poly}(n))$ time whether $\vx$ of length $n$ is in $L'$, where each $M_n$ can be constructed in $O(\mathrm{poly}(n))$ time as $M(\vtheta_n)$, for some Turing machine $M$ and some sequence of polynomially-sized \defn{advice strings} $\vTheta = \{ \vtheta_n \mid n \in \mathbb{N} \}$ with $|\vtheta_n| \in O(\mathrm{poly}(n))$. 
We define the language class $\mathrm{NP/poly}$ similarly to $\mathrm{P/poly}$: the only difference is the family $\{M_n: n \in \mathbb{N}\}$ consists of \emph{nondeterministic Turing machines}.

Naturally, $\mathrm{P} \subseteq \mathrm{P/poly}$. But $\mathrm{P/poly}$ is larger than $\mathrm{P}$: it contains all sparse languages, regardless of their hardness\Dash even sparse undecidable languages\Dash as well as many dense languages.  The extra power of $\mathrm{P/poly}$ comes from its access to compact advice strings that do not have to be recursively enumerable, let alone efficient to find.
This corresponds to statistical modeling, where the trained model has a computationally efficient architecture plus access to parameters that might have taken a long time to find.
\ifthenelse{\boolean{shortver}}{\looseness=-1}{%
}

\subsection{$\mathrm{NP}$-completeness and \textsc{Sat}}
\label{sec:sat-and-np}
$\mathrm{NP}$-complete decision problems have solutions that are efficient to validate but inefficient to find (assuming $\mathrm{P} \neq \mathrm{NP}$). 
One of the most well-known $\mathrm{NP}$-complete problems is the boolean satisfiability problem (\textsc{Sat}) \cite{cook1971}. Given a %
boolean formula $\phi$, $\textsc{Sat}$ accepts $\phi$ iff $\phi$ can be satisfied by some value assignment. For example, the formula $(A_1 \lor \neg A_2 \lor A_3) \land (A_1 \lor \neg A_4)$ is in $\textsc{Sat}$, since there is a satisfying assignment $A_{1\ldots 4} = \texttt{1101}$.
We denote the number of satisfying assignments to $\phi$ as $\#(\phi)$.%

It is widely believed that no $\mathrm{NP}$-complete languages are in $\mathrm{P/poly}$. Otherwise we would have all of $\mathrm{NP} \subseteq \mathrm{P/poly}$ 
and the polynomial hierarchy would collapse at the second level \cite{Karp1980SomeCB}.  

A capacity limitation of EC/ECCP weighted languages naturally follows from this belief:\footnote{All omitted proofs are in \cref{sec:proofs}.}
\begin{restatable}{lemma}{eccplimit}
	For any $L \in \mathrm{P}$, there exists an EC weighted language with support $L$.  For any $L \in \mathrm{P/poly}$, there exists an ECCP language with support $L$.  But for any $L \in \mathrm{NP\text{-}complete}$, there exists no ECCP language with support $L$ (assuming $\mathrm{NP}\nsubseteq\mathrm{P/poly}$).
	\label{thm:eccplimit}
\end{restatable}

In addition to not capturing the support of NP-complete languages, ECCP weighted languages cannot help solve other NP-hard  problems, either. 
For example, many structured prediction problems in NLP can be formulated as $\argmax_{\vx: \vxhat \prefixof \vx} \ptilde(\vx)$: we are given a prefix $\vxhat$ as input and look for its optimal continuation under $\tildep$.  But if this problem is $\mathrm{NP}$-hard for a particular $\ptilde$, then it is not in $\mathrm{P/poly}$ (assuming  $\npppoly$), so it cannot be accomplished by any polytime algorithm that queries an ECCP model.

%% file: s3.tex
\newcommand{\zero}{\mbox{\tt 0}}
\newcommand{\one}{\mbox{\tt 1}}
\newcommand{\two}{\mbox{\tt 2}}

\section{Autoregressive ECCP models (ELNCP models) have reduced capacity}
\label{sec:sat-reduction}
In this section we formally define \emph{autoregressive} ECCP models, and prove that they have strictly less capacity than general ECCP models or even just EC models.
Our proofs rely on the construction of a EC model $\ptilde$ where computing the local conditional probabilities $p(x \mid \hat{\vx})$ is $\mathrm{NP}$-hard, so they cannot be computed with compact parameters, if $\mathrm{NP}\nsubseteq\mathrm{P/poly}$.

\subsection{ELN and ELNCP models}
\label{sec:local-normalization}

Many parameter estimation techniques and inference methods specifically work with local conditional probabilities $p(x \mid \vxhat)$.  Thus, it is common to use parametric models where such quantities can be computed in time $O(\mathrm{poly}(|\vxhat|))$ (given the parameters).%
\footnote{An autoregressive model architecture  generally defines $p(\vx)$ as an efficiently computable (\cref{sec:computation-for-weighted-languages}) product of local conditional probabilities.  However, the parametrization usually ensures only that $\sum_{x \in V} p_{\vtheta}(x \mid \vxhat) = 1$ for all prefixes $\vxhat$.  Some parameter settings may give rise to \defn{inconsistent} distributions where $Z \defeq \sum_{\vx \in V^*} p_{\vtheta}(\vx) < 1$ because the generative process terminates with probability $< 1$ \cite{chen-etal-2018-recurrent}.  In this case, the factors $p_{\vtheta}(x \mid \vxhat)$ defined by the autoregressive model are not actually the conditional probabilities of the weighted language (as defined by \cref{sec:weighted-languages}). 
	 It is true that training $\vtheta$ with a likelihood objective does encourage finding a weighted language whose generative process always terminates (hence $Z=1$), since this is the behavior observed in the training corpus \cite{chi-geman-1998,chen-etal-2018-recurrent,welleck-etal-2020-consistency}.  
	 Our definitions of ELN(CP) models require the \emph{actual} conditional probabilities to be efficiently computable. 
	Autoregressive models that do not sum to $1$, whose normalized probabilities can be uncomputable, are not ruled out by our theorems that concern ELN(CP).
	\label{fn:consistency}}
These are the ``standard autoregressive models'' we discussed in \cref{sec:intro}.  We say that the resulting distributions are \defn{efficiently locally normalizable}, or \defn{ELN}.

We may again generalize ELNs to allow the use of compact parameters.  For any weighted language $\ptilde$, the Turing machine $\Mq$  \defn{efficiently locally normalizes $\ptilde$ with compact parameters} $\vThetaq = \{ \vthetaqn \mid n \in \mathbb{N} \}$ if\ifthenelse{\boolean{shortver}}{\looseness=-1}{%
}
\begin{itemize}[leftmargin=*]
    \item the parameter size $|\vthetaqn|$ grows only as  $O(\mathrm{poly}(n))$
	\item $\Mq(\vthetaq_n)$ returns a Turing machine $q_n$ (similar to $\tilde{p}_n$ in \cref{sec:compact}) in time $O(\mathrm{poly}(n))$ 
	\item $\ptilde$ is normalizable (so $p$ exists)
	\item 
	$q_n$ maps $\vxhat x \mapsto p(x \mid \vxhat)$ for all $x \in V \cup \{\eos\}$ and all prefixes $\vxhat \in V^*$ with $|\vxhat| \leq n$ and $Z(\vxhat) > 0$
	\item $q_n$ runs on those inputs $\vxhat x$ in time $O(\mathrm{poly}(n))$
\end{itemize}
If there is $\Mq$  that efficiently locally normalizes a weighted language $\ptilde$ with compact parameters $\vThetaq$, we say $\ptilde$ is \defn{efficiently locally normalizable with compact parameters}, or \defn{ELNCP}.  Note that this is a property of the weighted language itself.
\ifthenelse{\boolean{shortver}}{%
	In this case, it is obvious that $\ptilde$ %
	is ECCP:
}{%
	ELNCPs are more powerful than ELNs because their parameters may be set by an oracle:
	\begin{restatable*}{lemma}{inecinelncpnotineln}
		The set $\{\ptilde: \ptilde \in \mathrm{EC}, \ptilde \in \mathrm{ELNCP}, \ptilde \not\in\mathrm{ELN} \}$ is not empty.
		\label{thm:in-ec-in-elncp-not-in-eln}
	\end{restatable*}

It is also obvious that an ELNCP model %
is also ECCP:
}  
\begin{restatable}{lemma}{closure}
	\label{thm:closure}
	\label{fn:closure}
	An ELNCP model $\ptilde$ is also ECCP. Likewise, an ELN model is also EC. 
      \end{restatable}

If we define ELNCP \emph{models} analogously to ECCP models, \cref{thm:closure} means that locally normalized models do not provide any extra power.  Their distributions can always be captured by globally normalized models (of an appropriate architecture that we used in the proof). 
But we will see in \cref{thm:blowup} that 
the converse is likely not true: provided that $\mathrm{NP} \nsubseteq \mathrm{P/poly}$, there are efficiently computable weighted languages that cannot be efficiently locally normalized, even with the help of compact parameters.  That is, they are EC (hence ECCP), yet they are not ELNCP (hence not ELN).\ifthenelse{\boolean{shortver}}{\looseness=-1}{%
}

\subsection{ELNCP models cannot exactly capture all EC (or ECCP) distributions}
\label{sec:exact}
\ifthenelse{\boolean{shortver}}{%
	
}{%
\Cref{thm:in-ec-in-elncp-not-in-eln} states that with the help of compact parameters, ELNCP models can have some languages $\in \mathrm{P}$ as their supports which cannot be supports of ELN models. However, our proof of \cref{thm:in-ec-in-elncp-not-in-eln} depends on the observation that compact parameter vectors can encode any sparse language. Yet it is known that sparse $\mathrm{NP}$-complete languages exist if and only if $\mathrm{P}=\mathrm{NP}$ \cite{MAHANEY1982130}, which implies that assuming $\mathrm{P}\neq\mathrm{NP}$, we can no longer use the `sparsification' technique in the proof of \cref{thm:in-ec-in-elncp-not-in-eln} to model certificates of any $\mathrm{NP}$-complete language with ELNCP distributions. However, all such certificates are verifiable in polynomial time by definition (and hence model-able by EC distributions). Indeed, if we further assume $\npppoly$ (\cref{sec:sat-and-np}) \Dash an assumption even stronger than $\mathrm{P}\neq\mathrm{NP}$, yet still believed to be true among many \Dash we can show there exists a distribution that is $\mathrm{EC}$ but not $\mathrm{ELNCP}$.
}
\ifthenelse{\boolean{shortver}}{We reducing \textsc{Sat} to computing certain local conditional probabilities of $\tilde{p}$ (as defined in \cref{sec:weighted-languages}).    }{We prove our claim by defining a certain weighted language $\tilde{p}$ and reducing \textsc{Sat} to computing certain local conditional probabilities of $\tilde{p}$ (as defined in \cref{sec:weighted-languages}).  }
Each decision $\textsc{Sat}(\phi)$ (where $\phi$ ranges over formulas) corresponds to a particular local conditional probability, implying that there is no polytime scheme for computing all of these probabilities, even with polynomially sized advice strings (\emph{i.e.}, parameters).

Without loss of generality, we consider only formulae $\phi$ such that the set of variables mentioned at least once in $\phi$ is $\{A_1,\ldots,A_j\}$ for some $j\in\mathbb{N}$; we use $|\phi|$ to denote the number of variables $j$ in $\phi$.%
We say that $\va$ \defn{satisfies} $\phi$ if $\va \in \mathbb{B}^{|\phi|}$ and $(A_1=a_1,\ldots,A_{|\phi|}=a_{|\phi|})$ is a satisfying assignment.
Finally, let boldface $\vphi \in \mathbb{B}^*$ denote $\mathrm{enc}(\phi)$ where $\mathrm{enc}$ is a prefix-free encoding function.  
We can now define the unweighted language $\Xc = \{ \vphi\va \mid \phi \text{ is a formula and } \va \in \mathbb{B}^{|\phi|} \text{ and } \va \text{ satisfies } \phi\}$ over alphabet $\mathbb{B}$, which contains each possible \textsc{Sat} problem concatenated to each of its solutions.%
\footnote{For example, $\Xc$ contains the string $\vphi\va$ where $\vphi = \mathrm{enc}( (A_1\lor\neg A_2\lor A_3) \land (A_1 \lor \neg A_4 ) )$ and $\va=\mbox{\texttt{1101}}$. \label{ft:sat-example}}

We now convert $\Xc$ to a weighted language $\tilde{p}$, defined by $\tilde{p}(\vx) = \tilde{p}(\vphi, \va) =  (\frac{1}{3})^{|\vx|+1} $ for $\vx \in \Xc$ (otherwise $\tilde{p}(\vx)=0$).  $\tilde{p}$ is normalizable since $Z$ is both finite ($Z = \sum_{\vx \in \Bool^*} \tilde{p}(\vx) \leq \sum_{\vx \in \Bool^*}  (\frac{1}{3})^{|\vx|+1} = 1$) and positive ($Z > 0$ because the example string in \cref{ft:sat-example} has weight $> 0$).
The conditional distribution $p(\va \mid \vphi)$ is uniform over the satisfying assignments $\va$ of $\vphi$, as they all have the same length $|\phi|$.\ifthenelse{\boolean{shortver}}{\looseness=-1}{%
}

$\tilde{p}$ is efficiently computable, and so is $p =\tilde{p}/Z$.\footnote{Almost.  This $Z$ could be irrational, but at least it is computable to any desired precision.  
	For any rational $\hat{Z} \approx Z$, we can say $\hat{p} = \tilde{p}/\hat{Z} \approx p$ is EC, via a Turing machine $M^{\hat{p}}$ that stores $\hat{Z}$. Further remarks on irrationality 
	appear 
	in \cref{sec:proofs}.}  Yet deciding whether the local conditional probabilities of $\tilde{p}$ are greater than $0$ is $\mathrm{NP}$-hard. %
In particular, we show that \textsc{Sat}
 can be reduced to deciding whether certain local probabilities are greater than $0$, namely the ones that condition on prefixes $\hat{\vx}$ that consist only of a formula: $\hat{\vx}=\vphi$ for some $\phi$. This implies,  assuming $\mathrm{NP}\nsubseteq\mathrm{P/poly}$, that no $(\Mq, \vThetaq)$ can efficiently locally normalize $\tilde{p}$ with compact parameters. Granted, the restriction of $\tilde{p}$ to the finite set $\{\vx \in \mathbb{B}^*: |\vx| \leq n\}$ can be locally normalized by some polytime Turing machine $q_n$, using the same trie trick sketched in \cref{sec:compact}. But such tries have sizes growing exponentially in $n$, and it is not possible to produce a sequence of such machines, $\{q_n: n \in \mathbb{N}\}$, via a single master Turing machine $\Mq$ that runs in $O(\mathrm{poly}(n))$ on $\vthetaqn$. That is:%
\begin{restatable}{theorem}{blowup}
\label{thm:blowup} Assuming $\mathrm{NP} \nsubseteq \mathrm{P/poly}$, there exists an efficiently computable normalizable weighted language $\tilde{p}$ that is not ELNCP.
\end{restatable}%
\begin{proof}[Proof sketch]
Take $\ptilde$ to be the weighted language we defined earlier in this section.  $\ptilde$ is clearly efficiently computable.  We will show that if it is ELNCP via $(\Mq,\vThetaq)$, then the $\mathrm{NP}$-complete problem \textsc{Sat} is in $\mathrm{P/poly}$, contradicting the assumption.  We must give a method for using $(\Mq,\vThetaq)$ to decide $\textsc{Sat}$ in polytime and with compact parameters $\vTheta$.  Given $\phi$, our method constructs a simple related formula $\phi'$
such that
\begin{itemize}
    \item $\phi'$ has at least one satisfying assignment (so $Z(\vphi') > 0$ and thus $p(\one \mid \vphi')$ is defined)
    \item $\phi'$ has satisfying assignments with $A_1=\one$ (\emph{i.e.}, $p(\one \mid \vphi') > 0$) if and only if $\phi$ is satisfiable
\end{itemize}
Our construction also provides a polynomial function $f$ such that $|\vphi'|$ is guaranteed to be $\leq f(|\vphi|)$.%
We now define $\vTheta$ by $\vtheta_n = \vthetaq_{f(n)}$ $(\forall n)$.  When our \textsc{Sat} algorithm with compact parameters $\vTheta$ is given $\vphi$ of length $n$, it can use the polynomial-size advice string $\vtheta_n$ to ask $(\Mq,\vThetaq)$ in polynomial time for $p(\one \mid \vphi')$.  $\textsc{Sat}(\vphi)$ returns true iff that probability is $> 0$.\footnote{See also the remark on implications for seq2seq models following the proof in \cref{sec:proofs}.}
\end{proof}
\subsection{ELNCP models cannot even capture all EC (or ECCP) supports or rankings}

We can strengthen \cref{thm:blowup} as follows:

\begin{restatable}{theorem}{no-p-for-u} Assuming $\mathrm{NP}\nsubseteq\mathrm{P/poly}$,
	there exists an efficiently computable normalizable weighted language $\ptilde$ where there is no ELNCP $\qtilde$ such that $\mathrm{support}(\ptilde)=\mathrm{support}(\qtilde)$.
	\label{thm:nopforu}
\end{restatable}
\begin{proof}
	Observe that for any two weighted languages $\ptilde$ and $\qtilde$ with the same support, $\forall \vxhat \in V^*, Z_{\ptilde}(\vxhat) > 0 \iff Z_{\qtilde}(\vxhat) > 0$ (where $Z_{\ptilde}$ and $Z_{\qtilde}$ return the prefix probabilities of ${\ptilde}$ and $\qtilde$ respectively).  Thus, for any $\vxhat$ with $Z_{\ptilde}(\vxhat) > 0$, $p(\one \mid \vxhat) \defeq Z_{\ptilde}(\vxhat\one)/Z_{\ptilde}(\vxhat)$ and $q(\one \mid \vxhat) \defeq Z_{\qtilde}(\vxhat\one)/Z_{\qtilde}(\vxhat)$ are well-defined and $p(\one \mid \vxhat) > 0 \iff q(\one \mid \vxhat) > 0$.  If $\qtilde$ is ELNCP, then all such probabilities $q(\one \mid \vxhat)$ can be computed in polytime with compact parameters,
	so it is likewise efficient to determine whether $p(\one \mid \vxhat)  > 0$.  But this cannot be the case when $\ptilde$ is the weighted language used in the proof of \cref{thm:blowup}, since that would suffice to establish that $\textsc{Sat} \in \mathrm{P/poly}$, following the proof of that theorem.
\end{proof}

To put this another way, there exists an unweighted language in $\mathrm{P}$ (namely $\mathrm{support}(\ptilde)$)
that is not the support of \emph{any} ELNCP distribution.

If they have different support, normalizable languages also differ in their ranking of strings:
\begin{restatable}{lemma}{noequalranking}\label{thm:noequalranking}
	Let $\ptilde, \qtilde$ be normalizable weighted languages with $\mathrm{support}(\ptilde)\neq\mathrm{support}(\qtilde)$. Then
	$\exists \vx_1,$ $\vx_2 \in V^* $ such that $\ptilde(\vx_1) < \ptilde(\vx_2)$ but $\qtilde(\vx_1) \geq \qtilde(\vx_2)$.
\end{restatable}

Therefore, no ELNCP $\qtilde$ captures the string ranking of $\ptilde$ from \cref{thm:nopforu}. And for some $\ptilde$, any ELNCP $\qtilde$ misranks even string pairs of ``similar'' lengths:
\begin{restatable}{theorem}{noranking}\label{thm:noranking}
	Assuming $\mathrm{NP} \nsubseteq \mathrm{P/poly}$, there exists an efficiently computable normalizable weighted language $\ptilde$ such that no ELNCP $\qtilde$ with $\mathrm{support}(\qtilde) \supseteq \mathrm{support}(\ptilde)$  has $\ptilde(\vx_1) < \ptilde(\vx_2) \Rightarrow \qtilde(\vx_1) < \qtilde(\vx_2)$ for all $\vx_1, \vx_2 \in V^*$.
\
    Indeed, any such $\qtilde$ has a counterexample where $\ptilde(\vx_1)=0$.  Moreover, there is a polynomial
    $f_{\qtilde}:  \mathbb{N} \to \mathbb{N}$ such that a counterexample exists for {\em every} $\vx_1$ such that $\ptilde(\vx_1)=0$ and $\qtilde(\vx_1) > 0$, where the $\vx_2$ in this counterexample always satisfies $|\vx_2| \leq f_{\qtilde}(|\vx_1|)$. 
\end{restatable}
\Cref{thm:noranking} is relevant if one wishes to train a model $\qtilde$ to rerank strings that are proposed by another method (\emph{e.g.}, beam search on $\qtilde$, or exact $k$-best decoding from a more tractable distribution).  If the desired rankings are given by \cref{thm:noranking}'s $\ptilde$, any smoothed\footnote{Smoothing is used to avoid ever incorrectly predicting 0 (a ``false negative'') by ensuring $\mathrm{support}(\qtilde) \supseteq \mathrm{support}(\ptilde)$.  E.g., autoregressive language models often define $q(x \mid \vxhat)$ using a softmax over $V \cup \{\eos\}$, ensuring that $q(\vx) > 0$ for all $\vx\in V^*$.}
ELNCP model $\qtilde$ will misrank some sets of candidate strings, even sets all of whose strings are ``close'' in length, by failing to rank an impossible string ($\vx_1$ with $\ptilde(\vx_1)=0$) below a possible one ($\vx_2$ with $\ptilde(\vx_2)>0$).\ifthenelse{\boolean{shortver}}{\looseness=-1}{%
}

\subsection[whatever]{ELNCP models cannot even \emph{approximate} EC (or ECCP) distributions}
\label{sec:approximation-is-np-hard}

\Cref{thm:nopforu} implies that there exists $\tilde{p}$ whose local probabilities $p(x \mid \vxhat)$ are not approximated by any ELNCP $q$ to within any constant factor $\lambda$, since that would perfectly distinguish zeroes from non-zeroes and the resulting support sets would be equal.\footnote{%
Dropping the normalization requirement on the approximated local probabilities (so that possibly $\sum_{x \in V} q(x \mid \vxhat) \neq 1$) does not help. Otherwise, again, \textsc{Sat} could be solved in polynomial time (with the help of polysize advice strings) by using $q(\one \mid \vphi')$ to determine in the proof of \cref{thm:blowup} whether $p(\one \mid \vphi') > 0$.}

However, this demonstration hinges on the difficulty of multiplicative approximation of zeroes \Dash whereas real-world distributions may lack zeroes.
Below we further show that it is hard even to approximate the \emph{non-zero} local conditional probabilities (even with the additional help of randomness).
\label{sec:no-structural-zero-approx}
\begin{restatable}{theorem}{noapproxfullsupport}
Assuming $\mathrm{NP} \nsubseteq \mathrm{P/poly}$, there exists an efficiently computable weighted language $\ptilde : V^* \rightarrow \mathbb{R}_{\geq 0}$ such that there is no $(\Mq, \vThetaq)$ where $\vThetaq = \{ \vthetaqn \mid n \in \mathbb{N} \}$ that satisfies all of the following properties (similar to \cref{sec:local-normalization}):
\begin{itemize}[leftmargin=*]
\item the parameter size $|\vthetaq_n|$ grows only as $O(\mathrm{poly}(n))$
\item $\Mq(\vthetaq_n)$ returns a \emph{probabilistic} Turing machine $q_n$ in time $O(\mathrm{poly}(n))$
\item there exists $\lambda \geq 1$ such that 
for each $x \in V \cup \{\$\}$ and $\vxhat \in V^*$ with $|\vxhat| \leq n$ and $p(x \mid \vxhat) > 0$,
the probabilistic computation $q_n(\vxhat x)$ has probability $> \nicefrac{2}{3}$ of approximating $p(x \mid \vxhat)$ to within a factor of $\lambda$ (that is, $q_n(\vxhat x) / p(x \mid \vxhat) \in [1/\lambda,\lambda]$)
\item $q_n$ runs on those inputs $\vxhat x$ in time $O(\mathrm{poly}(n))$
\end{itemize}
Moreover, the statement above remains true
\begin{enumerate}
    \item[(a)] when the approximation guarantee is only required to hold for prefixes $\vxhat$ where $\{\vx: \vxhat \prefixof \vx\}$ is finite (so $p(x \mid \vxhat)$ is computable by brute force)\looseness=-1
    \item[(b)] or, when $\mathrm{support}(\ptilde) = V^*$
    \end{enumerate}
\label{thm:noapproxfullsupport}
\end{restatable}

\subsection{ELN models are \emph{unconditionally} weak}

Our above results rely on the \emph{$\mathrm{NP}$-hardness} of computing or approximating an EC distribution's autoregressive factors $p(\cdot \mid \vx_{<t})$.  In \cref{sec:proofs}, we show that these factors can even be \emph{uncomputable}.  In such cases, the distribution cannot be ELN (\cref{thm:in-ec-not-in-eln}), though sometimes it is still ELNCP (\cref{thm:in-ec-in-elncp-not-in-eln}).  This result does \emph{not} assume $\mathrm{P}\neq \mathrm{NP}$ or $\npppoly$.

%% file: sidestepping.tex
\section{Alternative model families}
\label{sec:sidestep}
We now discuss alternative families of sequence distributions that trade away efficiency or compactness in exchange for greater capacity, as shown in \cref{fig:modelfamilycomparison}.

\input{rlm}
\subsection{Latent-variable models}
\label{sec:marginalized}

Autoregressive models have $Z=1$ for any setting of the parameters (or at least any setting that guarantees consistency: see \cref{fn:consistency}).  Clearly $Z=1$ ensures that $Z$ is both finite and tractable.  Can we find a model family that retains this convenience (unlike EBMs), while still being expressive enough to have any non-empty language in $\mathrm{P}$ as support?

Autoregressive \emph{latent-variable} models form such a family.  As in directed graphical models, the use of latent variables provides a natural way to model partial observations of an underlying stochastic sequence of events. We will model an observed sequence $\vx$ of length $n$ as a function of a latent string $\vz$ of length $O(\mathrm{poly}(n))$.  As in EBMs, the probability $p(\vx)$ can be computationally intractable, allowing these models to break the expressivity bottleneck of ordinary autoregressive models.  However, the intractability no longer comes from exponentially many summands in the denominator $Z$, but rather from exponentially many summands in the numerator \Dash namely, the summation over all latent $\vz$ that could have produced $\vx$.  Notice that as a result, even unnormalized string weights are now hard to compute, although once computed they are already normalized.

 Formally, we define \defn{marginalized} weighted languages.
We say that $\ptilde$ is a \defn{marginalization} of the weighted language $\rtilde$ if it can be expressed as $\ptilde(\vx) = \sum_{\vz: \mu(\vz) = \vx} \rtilde(\vz)$, where $\mu: \Zc \rightarrow V^*$ is some function (the \defn{marginalization operator}).  We say it is a \defn{light marginalization} if $|\vz| \in O(\mathrm{poly}(|\mu(\vz)|))$ and $\mu$ runs in time $O(\mathrm{poly}(|\vz|))$.\footnote{WLOG, $\mu$ can be required to run in linear time $O(|\vz|)$, as it does in our constructions below.}  Typically $\mu(\vz)$ extracts a subsequence of $\vz$; it can be regarded as keeping the observed symbols while throwing away a polynomially bounded number of latent symbols.  

Light marginalizations of ELN distributions are a reasonable formalization of latent-variable autoregressive models.  They are more powerful than ELN distributions, and even include some distributions that (by \cref{thm:eccplimit}) are not even ELNCP or ECCP:
\setcounter{theorem}{6}
\begin{restatable}{theorem}{evalmarginalized}
\label{thm:evalmarginalized}
There exists a light marginalization $p$ of an ELN distribution, such that $\mathrm{support}(p)$ is an $\mathrm{NP}$-complete language. %
\end{restatable}

Our proof of \cref{thm:evalmarginalized} relies on special structure of a certain $\mathrm{NP}$-complete language (\textsc{Sat}) and does not evidently generalize to all languages in $\mathrm{NP}$.  

However, light marginalizations of \emph{ELNCP} distributions are more powerful still,%
\footnote{The capacity established by \cref{thm:nppolyifflmeccp} does not need the full power of marginalization.  We could similarly define light \emph{maximizations} of ELNCP distributions, $\ptilde(\vx) = \max_{\vz: \mu(\vz)=\vx} \rtilde(\vz)$.  Replacing sum by max does not change the support.}
and can have \emph{any} language $\in\mathrm{NP}$ or even $\mathrm{NP/poly}$ (\cref{sec:p-poly}) as support:
\begin{restatable}{theorem}{nppolyifflmeccp}
	The following statements are equivalent for any nonempty $L \subseteq V^*$:
	\begin{enumerate}[(a)]
		\item $L \in \mathrm{NP/poly}$.
		\item $L$ is the support of a light marginalization of an ELNCP distribution.
		\item $L$ is the support of a light marginalization of an ECCP weighted language.
	\end{enumerate}
\label{thm:nppolyifflmeccp}
\end{restatable}

\Cref{thm:evalmarginalized,thm:nppolyifflmeccp} make use of unrestricted latent-variable autoregressive models.  There exist more practical restricted families of such models that admit tractable computation of $p(\vx)$  \cite{lafferty2001,rastogi-etal-2016-weighting,Wu2018HardNA,buys-blunsom-2018-neural}.  Such models are EC (and indeed, typically ELN) \Dash but this limits their expressivity, by \cref{thm:blowup}. Both \citet{lin-etal-2019-neural} and \citet{buys-blunsom-2018-neural} observed that such models yield worse empirical results than models that do not have tractable exact inference methods.  The tractability requirement is dropped in ``self-talk'' \cite{gpt3comment,Gontier2020MeasuringSG,shwartz-etal-2020-unsupervised}, where a neural autoregressive language model generates an analysis of the prefix $\vxhat$ via latent intermediate symbols before predicting the next output symbol.%
\footnote{Here the marginal distribution of the next observed symbol can require superpolynomial time to compute (if  $\mathrm{\#P} \neq \mathrm{FP}$, which follows from $\npppoly$).  \Cref{thm:blowup} could likewise be evaded by \emph{other} autoregressive approaches that invest superpolynomial computation in predicting the next symbol \cite{Graves2016AdaptiveCT}.  Each autoregressive step might explicitly invoke lookahead or reasoning algorithms, just as feed-forward network layers can invoke optimizers or solvers \cite{amos2017optnet,Wang2019SATNetBD}.\ifthenelse{\boolean{shortver}}{\looseness=-1}{%
}}

We remark that for autoregressive models, the \emph{position} of the latent variables is significant.  Marginalizing out latent variables at the \emph{end} of the string adds no power.  More precisely, if an ELNCP distribution is over strings $\vz$ of the form $\vx\#\vy$, then its marginalization via $\mu(\vx\#\vy) = \vx$ can be expressed more simply as an ELNCP language.  
Thus, by \cref{thm:nopforu}, marginalizations of such distributions cannot have arbitrary $\mathrm{NP}$ languages as support.
Our proofs of \cref{thm:evalmarginalized,thm:nppolyifflmeccp} instead use latent strings of the form $\vy\#\vx$, where all latent variables precede all observed ones \citep[as in][]{Kingma2014AutoEncodingVB}. 
(This simple design can always be used without loss of generality.) 
Trying to reorder those latent strings as $\vx\#\vy$ while preserving their weights would have yielded a non-ELNCP distribution $p(\vx\#\vy)$ (because if it were ELNCP, then $p(\vx)$ would be ELNCP also, and we know from \cref{thm:eccplimit} that it cannot be for any distribution whose support is an $\mathrm{NP}$-complete language).

How about lightly marginalizing ECCP languages instead of ELNCP ones? This cannot model any additional \emph{unweighted} languages, by \cref{thm:nppolyifflmeccp}.  But it may be able to model more probability distributions.  One can easily construct a light marginalization $p$ of an ECCP distribution such that $\#(\phi) = c_n \cdot p(\vphi)$, where $\#(\phi)$ is the number of satisfying assignments of $\phi$ and the constant $c_n$ depends only on $n=|\vphi|$.  We conjecture that this is not possible with lightly marginalized ELNCP distributions.

\subsection{Lookup models}
\label{sec:semiparametric-models}

\Cref{sec:compact} noted that with exponential growth in stored parameters, it is possible to fit any weighted language up to length $n$, with local probabilities computed in only $O(n)$ time by lookup.  Of course this rapidly becomes impractical as $n$ increases, even if the amount of training data increases accordingly.  However, there has been some recent movement toward storage-heavy models.
Such models are typically semiparametric: they use a parametric neural model, such as an autoregressive model, together with an external 
knowledge base of text strings or factoids that are not memorized in the layer weights.  The neural model generates queries against the knowledge base and combines their results.  Examples include $k\mathrm{NN}$LMs \cite{khandelwal20generalization} and semiparametric LMs \cite{Yogatama2021AdaptiveSL}.  The knowledge base grows linearly with the training data rather than compressing the data into a smaller parameter vector.  It is in fact a copy of the training data, indexed to allow fast lookup \cite{indyk1998}. (Preparing the index is much cheaper than neural network training.)  Access to the large knowledge base may reduce the amount of computation needed to find the local conditional probabilities, much as in the trie construction of \cref{sec:compact}.

%% file: rlm.tex
\subsection{Energy-based models (EBMs)}
\label{sec:ebm}

\defn{Energy-based models} \cite{lecun-06} of discrete sequences \cite{rosenfeld2001,sandbank2008refining,huang2018} traditionally refer to the EC models of \cref{sec:computation-for-weighted-languages}.  Only the unnormalized probabilities $\ptilde_{\vtheta}(\vx)$ are required to be efficiently computable. \Cref{thm:closure,thm:eccplimit} showed that this model family contains all ELN languages and can achieve any support in $\mathrm{P}$.  \Cref{thm:blowup} shows that it also contains languages that are not ELN or even ELNCP: intuitively, the reason is that the sums $Z(\vxhat)$ needed to compute the local normalizing constants (see \cref{sec:weighted-languages}) can be intractable.

If we generalize energy-based sequence models to include all ECCP models\Dash that is, we allow non-uniform computation with compact parameters\Dash then \cref{thm:closure,thm:eccplimit} guarantee that they can capture all 
ELNCP languages and furthermore all languages 
in $\mathrm{P/poly}$ (though still not $\mathrm{NP}$-complete languages).

\paragraph{Experiments on different parameterizations.}
Maximum-likelihood parameter estimation (MLE) can be expensive in an
EBM because the likelihood formula involves the expensive summation
$Z = \sum_{\vx \in V^*} \ptilde_{\vtheta}(\vx)$.  This forces us in
practice to use alternative estimators that do not require computing
normalized probabilities, such as noise-contrastive estimation (NCE)
or score matching (\cref{sec:intro}), which are less statistically
efficient.  In pilot experiments we found that both RNN- and Transformer-based EBMs trained with NCE achieved \emph{worse} held-out perplexity than
comparable locally normalized models trained with MLE.\footnote{This might be due to a
capacity limitation of the specific globally normalized architectures
(\emph{i.e.}, no parameters work well), or excess capacity (\emph{i.e.}, too many parameters work well on the finite sample), or statistical inefficiency of the estimator (the NCE objective on the finite sample, with the noise distribution we chose, does not distinguish among parameters as well as MLE does), or an optimization difficulty caused by local optima in the NCE optimization landscape.}\ifthenelse{\boolean{shortver}}{\looseness=-1}{%
}

Fortunately, it is possible to infuse a globally normalized
architecture with the inductive bias of a locally normalized one, which empirically yields good results. \defn{Residual energy-based models} (\defn{REBMs})  \cite{deng2020residual} are a simple hybrid architecture:
\begin{align*}
    p_{\vtheta}(\vx) &\propto \tilde{p}_{\vtheta}(\vx) \triangleq p_0(\vx) \cdot \exp \g(\vx) \label{eq:residual-sum} 
\end{align*}
This simply multiplies our previous weight by a new factor $p_0(\vx)$.  The \emph{base model} $p_0: \Xc \rightarrow (0, 1]$
is a locally normalized neural sequence model (ELN model) that was pretrained on the same distribution.  %
$\g: V^* \rightarrow \mathbb{R}$
is a learnable function (with parameters $\vtheta$) that is used to adjust $p_0$, yielding a weighted language $\ptilde_\vtheta$ with the same support
$\Xc$.
We implemented REBMs, again with NCE training, and evaluated them on two different neural architectures (GRU- and Transformer-based) and $3$ datasets (WikiText \cite{Merity2017PointerSM}, Yelp \cite{yelp}, and RealNews \cite{zellers2019defending}). In each setting we tried, the REBM slightly but significantly improved the perplexity of the base model $p_0$  ($p < 0.05$).\footnote{We independently conceived of and implemented the REBM idea proposed in \citet{deng2020residual}.  Details of neural architecture choice, model parameter sizes, training regimen, and evaluation (\cref{sec:rlm-details,sec:comparison-local,sec:experimental-details}) differ between our work and theirs, which also reported positive empirical results (on different datasets). We regard the two independent positive findings as a strong indication that the REBM design is effective.
}

%% file: related.tex
\section{Related work}%
\Citet{chen-etal-2018-recurrent} show that it is hard to map \emph{RNN parameters} to properties of the resulting autoregressive weighted language, such as consistency ($Z=1$).  We focus on cases where the RNN parameters are already known to be consistent, so the RNN efficiently maps a \emph{string} $\vxhat$ to its local conditional distribution $p(\cdot \mid \vxhat)$.  Our point is that for some weighted languages, this is not possible (even allowing polynomially larger RNNs for longer strings), so consistent RNNs and their ilk cannot be used to describe such languages.

In a Bayes network\Dash which is really just an autoregressive model of fixed-length strings\Dash approximate marginal inference is $\mathrm{NP}$-hard \cite{roth1996}.
Assuming $\npppoly$ and the grid-minor hypothesis, \citet[Theorem 5.6]{Chandrasekaran2008} further showed that for any infinite sequence of graphs $G_1, G_2, \ldots$ where $G_n$ has treewidth $n$, there is no sequence of algorithms $M_1, M_2, \ldots$ such that $M_n$ performs approximate marginal inference in time $O(\mathrm{poly}(n))$ on graphical models of structure $G_n$.  This remarkable negative result says that in \emph{any} graph sequence of unbounded treewidth, approximating the normalizing constant for $G_n$ given arbitrary parameters is hard (not $O(\mathrm{poly}(n))$), even with advice strings.  Our negative result (\cref{thm:noapproxfullsupport}) focuses on \emph{one particular} infinite weighted language, showing that approximating local conditional probabilities given an arbitrary length-$n$ prefix is hard in the same way.  (So this language cannot be captured by an RNN, even with advice strings.)\ifthenelse{\boolean{shortver}}{\looseness=-1}{%
}

%% file: conclusion.tex
\section{Conclusion and future work}
\label{sec:conclusion}

Autoregressive models are suited to those probability
distributions whose prefix probabilities are efficiently computable.
This efficiency is convenient for training and sampling.
But unless we sacrifice it and allow runtime or parameter size to grow superpolynomially in input length,
autoregressive models are less expressive than models whose prefix probabilities expensively marginalize over suffixes or latent variables.

All model families we have discussed in this paper can be seen as making compromises between different desiderata (\cref{fig:modelfamilycomparison}). Natural follow-up questions include \emph{`Are there model families that win on all fronts?' `What are other modeling desiderata?'} 

While \emph{some} languages $\in \mathrm{P}$ cannot be supports of ELNCPs, we do not know if the same can be said for \emph{most} languages $\in \mathrm{P}$.
This problem seems to be closely related to the average complexity of $\mathrm{NP}$-complete languages, where most questions remain open \citep{Levin1986AverageCC,bogdanov2006}. 

%% file: appendix.tex
\begin{figure*}[ht!]
	\centering
	\includegraphics[width=1.\linewidth]{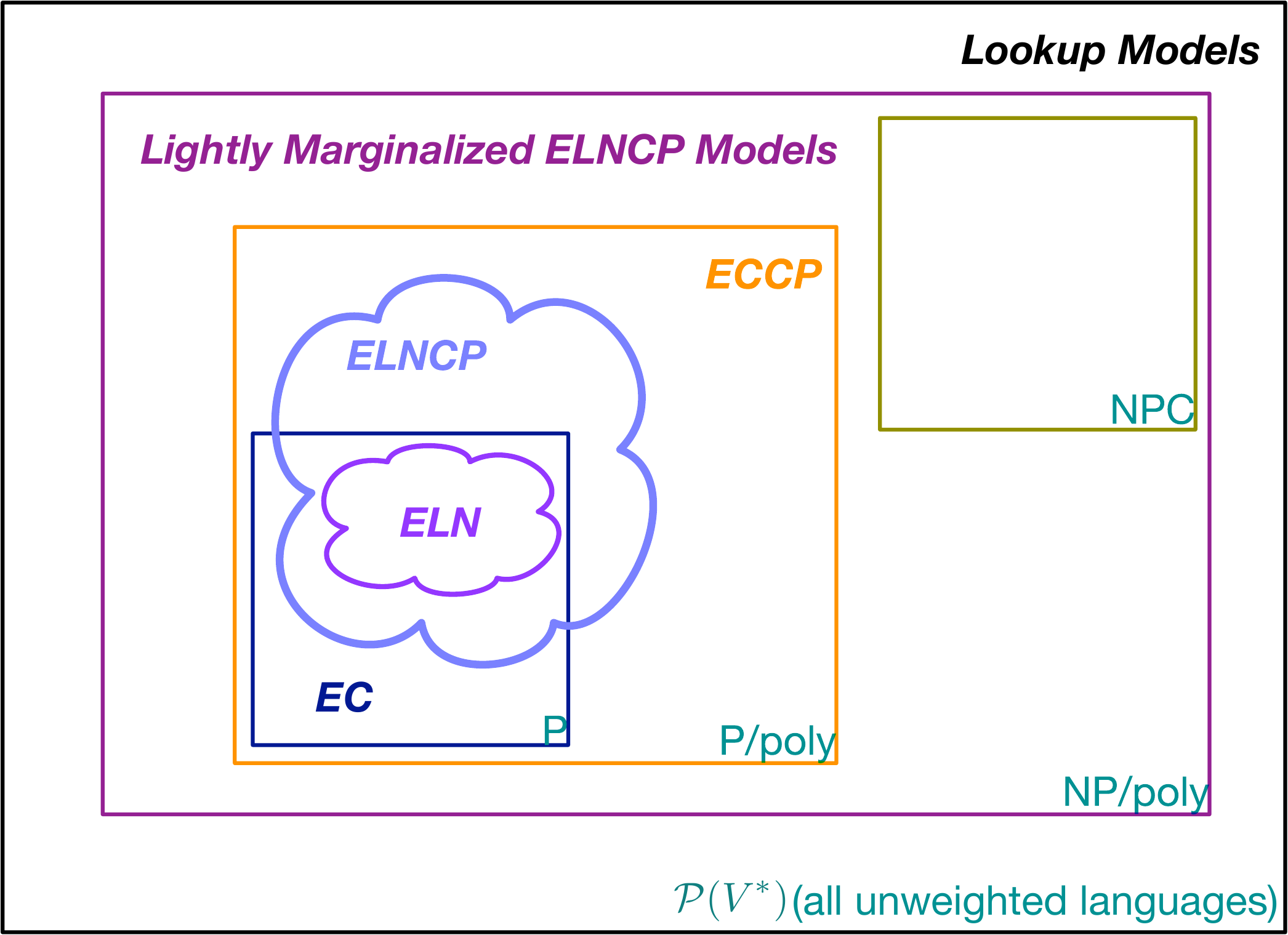}
	\caption{The space of unweighted languages.  We assume in this diagram that $\mathrm{NP}\nsubseteq\mathrm{P/poly}$.  
	Each rectangular outline corresponds to a complexity class (named in its lower right corner) and encloses the languages whose decision problems fall into that class. Each \textbf{\textit{bold-italic label}} (colored to match its shape outline) 
	names a model family and encloses the languages that can be expressed as the \emph{support} of some \emph{weighted} language in that family. All induced partitions in the figure are non-empty sets: shape A properly encloses shape B if and only if language class A is a strict superset of language class B.
	As mentioned in \cref{fig:modelfamilycomparison}, standard autoregressive models (ELN models) have support languages that form a strict subset of $\mathrm{P}$ (\cref{thm:eccplimit,thm:closure,thm:in-ec-not-in-eln,sec:p-poly}).
	ELNCP models (\cref{sec:local-normalization}) extend ELN models by allowing the parameter size to grow polynomially in string length, allowing them to capture both more languages inside $\mathrm{P}$ (\cref{thm:in-ec-in-elncp-not-in-eln}) and languages outside $\mathrm{P}$ (including undecidable but sparse languages) that can be characterized autoregressively with the help of these compact parameters.  All of those languages belong in the class $\mathrm{P/poly}$.  \Cref{thm:nopforu} establishes that energy-based (EC) and ECCP models go strictly further than ELN and ELNCP models, respectively (\cref{thm:nopforu}): they correspond to the \emph{entire} classes $\mathrm{P}$ and $\mathrm{P/poly}$ (\cref{thm:eccplimit}).
	However, even ECCP does not capture any $\mathrm{NP}$-complete languages under our assumption $\mathrm{NP}\nsubseteq\mathrm{P/poly}$.  
	Allowing a polynomial number of latent symbols extends the power further still: lightly marginalized ELNCP or ECCP distributions cover exactly the languages $\in \mathrm{NP/poly}$ (\cref{thm:nppolyifflmeccp}).
		Finally, if we were to drop the requirement that the parameters $\vTheta$ must be compact, we could store lookup tries to model any weighted language
		(\cref{sec:semiparametric-models}).
	}
	\label{fig:modelsupportzoo}
\end{figure*}
\clearpage
\section{Proofs}\label{sec:proofs}
\eccplimit*
This simple lemma relates our classes EC and ECCP of \emph{weighted} languages to the complexity classes $\mathrm{P}$ and $\mathrm{P/poly}$ of their supports, which are \emph{unweighted} formal languages (\cref{sec:background}).  It holds because computing a string's weight can be made as easy as determining whether that weight is nonzero (if we set the weights in a simple way), but is certainly no easier.  We spell out the trivial proof to help the reader gain familiarity with the formalism.
\begin{proof}
	Given $L$, define a weighted language $\ptilde$ with support $L$ by $\ptilde(\vx) = 1$ if $\vx \in L$ and $\ptilde(\vx) = 0$ otherwise.  
	
	If $L \in P$, then clearly $\ptilde$ is EC since the return value of 1 or 0 can be determined in polytime.  
	
	If $L \in \mathrm{P/poly}$, $L$ can be described as a tuple $(M, \vTheta)$ following our characterization in \cref{sec:p-poly}.  It is easy to show that $\ptilde$ is ECCP, using the same polynomially-sized advice strings $\vTheta$.  We simply construct $\Mtildep$ such that $\Mtildep(\vtheta_n)$ returns $1$ or $0$ on input $\vx$ according to whether $M(\vtheta_n)$ accepts or rejects $\vx$.  Both $\Mtildep(\vtheta_n)$ and $\Mtildep(\vtheta_n)(\vx)$ are computed in time $O(\mathrm{poly}(n))$ if $|\vx|=n$.  (The technical construction is that $\Mtildep$ simulates the operation of $M$ on the input $\vtheta_n$ to obtain the description of the Turing machine $M_n = M(\vtheta_n)$, and then outputs a slightly modified version of this description that will write $1$ or $0$ on an output tape.)
	
	For the second half of the lemma, we use the reverse construction.  Suppose $\ptilde$ is an ECCP weighted language with support $L$.  $\ptilde$ can be characterized by a tuple $(\Mtildep, \vTheta)$.  It is easy to show that $L \in \mathrm{P/poly}$, using the same polynomially-sized advice strings $\vTheta$.  We simply construct $M$ such that $M(\vtheta_n)$ accepts $\vx$ iff $\Mtildep(\vtheta_n)(\vx) > 0$.  Then by the assumption, $L \notin \mathrm{NP}$-complete.
\end{proof}

\closure*
\begin{proof}
Let $\ptilde$ be an ELNCP language.  This implies that $\ptilde$ is normalizable, so let $p(\vx) \defeq \ptilde(\vx)\,/\,Z$ as usual.
Specifically, let $\Mq$ efficiently locally normalize $\ptilde$ with compact parameters $\vThetaq = \{ \vthetaqn \mid n \in \mathbb{N} \}$.  It is simple to define a Turing machine $\Mr$ that maps each parameter string $\vthetaqn$ to a Turing machine $r_n$, where $r_n(\vx)$ simply computes $\left( \prod_{t=1}^{n} q_n( x_t \mid \vx_{<t} ) \right) \cdot q_n(\$ \mid \vx)$.  Then for all $\vx$ of length $n$, $r_n(\vx) = \left( \prod_{t=1}^{n} p( x_t \mid \vx_{<t} ) \right) \cdot p(\$ \mid \vx)$, by the definition of local normalization, and thus $r_n(\vx) = p(\vx)$.

$\Mr$ can be constructed by incorporating the definition of $\Mq$, so that $r_n = \Mr(\vthetaqn)$ can include $q_n = \Mq(\vthetaqn)$ as a subroutine.  This allows $r_n$ to query $q_n$ for local conditional probabilities and multiply them together.
\begin{itemize}[leftmargin=*]
\item Since $\Mq$ runs in polytime, it is straightforward for this construction to ensure that $\Mr$ runs in polytime as well.
\item Since $q_n(\cdot \mid \vxhat) \in O(\mathrm{poly}(n))$, this construction can ensure that $r_n$ runs in polytime as well.
\item We were given that $|\vthetaqn| \in O(\mathrm{poly}(n))$ (compact parameters).  
\end{itemize} 
Since $p$ is the weighted language defined by $(\Mr, \vThetaq)$, and $\Mr$ and $\vThetaq$ have the properties just discussed, we see that $p$ is efficiently computable with compact parameters (ECCP). Therefore $\ptilde(\vx) = Z p(\vx)$ is also ECCP.

In the case where $\ptilde$ is more strongly known to be ELN (the parameters $\vThetaq$ are not needed), a simplification of this argument shows that it is EC.
\end{proof}

\blowup*
\begin{proof}
The proof was sketched in \cref{sec:exact}.  Here we fill in the details.  

The unweighted language $\tilde{p}$ defined in that section is efficiently computable via the following simple algorithm that outputs $\tilde{p}(\vx)$ given $\vx \in \Bool^*$. If $\vx$ has a prefix that encodes a formula $\phi$, and the remainder of $\vx$ is a satisfying assignment $\va$ to the variables of $\phi$, then return $(\frac{1}{3})^{|\vx|+1}$.  Otherwise return 0.  This algorithm can be made to run in polynomial time because whether an assignment satisfies a formula can be determined in polynomial time (a fact that is standardly used to establish that $\textsc{Sat} \in \mathrm{NP}$).

Given a formula $\phi$ with variables $A_1, \ldots, A_j$, we define $\phi' = (\neg A_{1} \land \neg A_{2} \land \ldots \land \neg A_j \land \neg A_{j+1}) \lor (A_{1} \land \mathrm{Shift}(\phi))$, where $\mathrm{Shift}(\phi)$ is a version of $\phi$ in which $A_i$ has been renamed to $A_{i+1}$ for all $1 \leq i \leq j$.  
It is obvious that $\phi'$ and $p$ have the properties stated in the proof sketch.  The strings in $\Xc$ that begin with $\vphi'$ are precisely the strings of the form $\vphi'\va'$ where $\va'$ is a satisfying assignment of $\phi'$\Dash which happen just when $\va'=\zero^{j+1}$ or $\va'=\one \va$ where $\va$ is a satisfying assignment of $\phi$.  
At least one string in $\Xc$ begins with $\vphi'$, namely $\vphi'\zero^{j+1}$, so $Z(\vphi') > 0$. Moreover, $Z(\vphi' \one) >  0$ iff $\phi$ has any satisfying assignments.  Therefore the local probability $p(\one \mid \vphi') = Z(\vphi' \one)\,/\,Z(\vphi')$ is defined (see \cref{sec:weighted-languages}), and is $> 0$ iff $\textsc{Sat}(\phi)$.

Notice that the formal problem used in the proof is a version of $\textsc{Sat}$ whose inputs are encoded using the same prefix-free encoding function $\mathrm{enc}$ that was used by our definition of $\Xc$ in \cref{sec:exact}.  We must choose this encoding function to be concise in the sense that $\vphi \defeq \mathrm{enc}(\phi)$ can be converted to and from the conventional encoding of $\phi$ in polynomial time.  This ensures that our version of $\textsc{Sat}$ is $\leq^P_m$-interreducible with the conventional version and hence NP-complete.  It also ensures that there is a polynomial function $f$ such that $|\vphi'| \leq f(|\vphi|)$, as required by the proof sketch, since there is a polynomial-time function that maps $\vphi \to \phi \to \phi' \to \vphi'$ and the output length of this function is bounded by its runtime.  This is needed to show that our version of \textsc{Sat} is in $\mathrm{P/poly}$.

Specifically, to show that the existence of $(\Mq, \vThetaq)$ implies $\textsc{Sat} \in \mathrm{P/poly}$, we use it to construct an appropriate pair $(M,\vTheta)$ such that $(M(\vtheta_n))(\vphi) = \textsc{Sat}(\phi)$ if $|\vphi|=n$. As mentioned in the proof sketch, we define $\vTheta$ by $\vtheta_n = \vthetaq_{f(n)}$, and observe that $|\vtheta_n| \in O(\mathrm{poly}(n))$ (thanks to compactness of the parameters $\vThetaq$ and the fact that $f$ is polynomially bounded).  Finally, define $M(\vtheta_n)$ to be a Turing machine that maps its input $\vphi$ of length $n$ to $\vphi'$ of length $\leq f(n)$, then calls $\Mq(\vtheta_n) = \Mq(\vthetaq_{f(n)})$ on $\vphi' \one$ to obtain $p(\one \mid \vphi')$, and returns true or false according to whether $p(\one \mid \vphi') > 0$.  Computing $\vphi'$ takes time polynomial in $n$ (thanks to the properties of $\mathrm{enc}$). Constructing $\Mq(\vtheta_{f(n)})$ and calling it on $\vphi'$ each take time polynomial in $n$ (thanks to the properties of $f$ and $\Mq$).  
\end{proof}

\paragraph{Remark on  conditional models.}\label{para:seq2seqimplication} While we focus on modeling \emph{joint} sequence probabilities in this work, we note that in many applications it often suffices to just model \emph{conditional} probabilities \cite{10.5555/2969033.2969173}.%
Unfortunately, our proof of \cref{thm:blowup} above implies that ELNCPs do not make good conditional models either: specifically, there exists $\vphi$ such that deciding whether $p(\texttt{1} \mid \vphi) > 0$ is $\mathrm{NP}$-hard, and thus beyond ELNCP's capability.

\paragraph{Remark on irrationality.}\label{para:irrational} In our definitions of ECCP and ELNCP languages, we implicitly assumed that the Turing machines that return weights or probabilities would write them in full on the output tape, presumably as the ratio of two integers.  Such a Turing machine can only return rational numbers.

But then our formulation of \cref{thm:blowup} allows another proof.  We could construct $\ptilde$ such that the local conditional probabilities $p(x \mid \vxhat) \defeq Z(\vxhat x)/Z(\vxhat)$ are sometimes irrational.  In this case, they cannot be output exactly by a Turing machine, implying that $\ptilde$ is not ELNCP.  However, this proof exposes only a trivial weakness of ELNCPs, namely the fact that they can only define distributions whose local marginal probabilities are rational.

We can correct this weakness by formulating ELNCP languages slightly differently.  A real number is said to be \defn{computable} if it can be output by a Turing machine to any desired precision.  That Turing machine takes an extra input $b$ which specifies the number of bits of precision of the output.  Similarly, our definitions of ECCP and ELNCP can be modified so that their respective Turing machines $\ptilde_n$ and $q_n$ take this form, are allowed to run in time $O(\mathrm{poly}(n+b))$, and have access to the respective parameter vectors $\vThetap_{n+b}$ and $\vThetaq_{n+b}$.  Since some of our results concern the ability to distinguish zero from small values (arbitrarily small in the case of \cref{thm:in-ec-in-elncp-not-in-eln}), our modified definitions also require $\ptilde_n$ and $q_n$ to output a bit indicating whether the output is \emph{exactly} zero.  For simplicity, we suppressed these technical details from our exposition.

Relatedly, in \cref{sec:semiparametric-models}, we claimed that lookup models can fit any weighted 
language up to length $n$.  This is not strictly true if the weights can be irrational.  A more precise statement is that for any weighted language $\ptilde$, there is a lookup model that maps $(\vx, b)$ to the first $b$ bits of $\ptilde(\vx)$.  Indeed, this holds even when $\ptilde(\vx)$ is uncomputable.

\paragraph{Remark on computability.}\label{para:computability} In \cref{sec:weighted-languages} we claimed that any weighted language $\ptilde$ that has a finite and strictly positive $Z$ can be normalized as $p(\vx) = \nicefrac{\ptilde(\vx)}{Z}$. However, $Z$ may be uncomputable: that is, there is no algorithm that takes number of bits of precision $b$ as input, and outputs an approximation of $Z$ within $b$ bits of precision. 
Therefore, even if $\ptilde$ is computable, $p$ may have weights that are not merely irrational but even \emph{uncomputable}.  An example appears in the proof of \cref{thm:in-ec-in-elncp-not-in-eln} below.  Weighted language classes (\emph{e.g.} ELNCP) that only model normalized languages will not be able to model such languages, simply because the partition function is uncomputable. 

However, our proof of \cref{thm:blowup} does not rely on this issue, because the $\ptilde$ that it exhibits happens to have a computable $Z$.  For any $b$, $Z$ may be computed to $b$ bits of precision as the explicit sum $\sum_{\vx: |\vx|\leq N} \ptilde(\vx)$ for a certain large $N$ that depends on $b$.%

\paragraph{Remark on RNNs.}
Our proof of \cref{thm:blowup} showed that our problematic language $\tilde{p}$ is efficiently computable (though not by any locally normalized architecture with compact parameters).  Because this paper is in part a response to popular neural architectures, we now show that $\tilde{p}$ can in fact be computed efficiently \emph{by a recurrent neural network (RNN)} with compact parameters.  Thus, this is an example where a simple globally normalized RNN parameterization is fundamentally more efficient (in runtime or parameters) than any locally normalized parameterization of any architecture (RNN, Transformer, etc.).  

Since we showed that $\tilde{p}$ is efficiently computable, the existence of an RNN implementation is established in some sense by the ability of finite rational-weighted RNNs to simulate Turing machines \cite{Siegelmann1992OnTC}, as well as an extension to \citet[Thm. 11]{chen-etal-2018-recurrent} to a family of RNNs, where each RNN instance also takes some formula encoding as input. However, it is straightforward to give a concrete construction, for each $n \in \mathbb{N}$, for a simple RNN that maps each string $\vx \in \mathbb{B}^n$ to $\tilde{p}(\vx)$.  Here $\tilde{p}(\vx)$
will be either $(\frac{1}{3})^{n+1}$ or 0, according to whether $\vx$ has the form $\vphi\va$ where $\vphi$ encodes a \textsc{3-CNF-Sat} formula $\phi$ that is satisfied by $\va$.\footnote{The restriction to \textsc{3-CNF-Sat} formulas is convenient, but makes this a slightly different definition of $\Xc$ and $\ptilde$ than we used in the proofs above.  Those proofs can be adjusted to show that this $\ptilde$, too, cannot be efficiently locally normalized with compact parameters.  The only change is that in the construction of \cref{thm:blowup}, $\phi'$ must be converted to 3-CNF.  The proof then obtains its contradiction by showing that $\textsc{3-CNF-Sat} \in \mathrm{P/poly}$ (which suffices since \textsc{3-CNF-Sat} is also NP-complete).}
The basic idea is that $\vphi$ has $j \leq n$ variables, so there are only $O(n^3)$ possible 3-CNF clauses. The RNN allocates one hidden unit to each of these.  When reading $\vphi\va$, each clause encountered in $\vphi$ causes the corresponding hidden unit to turn on, and then each literal encountered in $\va$ turns off the hidden units for all clauses that would be satisfied by that literal.  If any hidden units remain on after $\vx$ has been fully read, then $\vphi$ was not satisfied by $\va$, and the RNN's final output unit should return 0.  Otherwise it should return $(\frac{1}{3})^{n+1}$, which is constant for this RNN.  To obtain digital behaviors such as turning hidden units on and off, it is most convenient to use ramp activation functions for the hidden units and the final output unit, rather than sigmoid activation functions.  Note that our use of a separate RNN $M^{\textrm{RNN}}_n$ for each input length $n$ is an example of using more hidden units for larger problems, a key idea that we introduced in \cref{sec:compact} in order to look at asymptotic behavior. The RNN's parameter sequence $\vTheta^{\textrm{RNN}} = \{ \vtheta^{\textrm{RNN}}_n \mid n \in \mathbb{N} \}$ is obviously compact, as $\vtheta^{\textrm{RNN}}_n$ only has to store the input length $n$. With our alphabet $\Bool$ for $\tilde{p}$, $|\vtheta^{\textrm{RNN}}_n| \in O(\log n)$.
\noequalranking*
\begin{proof}
    Suppose that the claim is false, \emph{i.e.}, $\ptilde$ and $\qtilde$ have the same ranking of strings.  Then the minimum-weight strings under $\ptilde$ must also be minimum-weight under $\qtilde$.  WLOG, there exists $\vx \in V^*$ with $\ptilde(\vx)=0$ and $\qtilde(\vx) = c > 0$.  Then $c > 0$ is the minimum weight of strings in $\qtilde$.  But this is not possible for a normalizable language $\qtilde$, since it means that $Z_{\qtilde} \defeq \sum_{\vx' \in V^*} q(\vx') \geq \sum_{\vx' \in V^*} c$ diverges. 
\end{proof}

\noranking*

\begin{proof}
Let $\ptilde$ be the weighted language from \cref{thm:nopforu}.  Given an ELNCP $\qtilde$.  By \cref{thm:nopforu}, $\mathrm{support}(\qtilde) \neq \mathrm{support}(\ptilde)$, so there must exist a string $\vx_1$ that is in one support language but not the other.  With the additional assumption that $\mathrm{support}(\qtilde) \supseteq \mathrm{support}(\ptilde)$, it must be that $\vx_1 \in \mathrm{support}(\qtilde)$, so $\ptilde(\vx_1)=0$ but $\qtilde(\vx_1) > 0$.  

Given \emph{any} such $\vx_1$ with $\ptilde(\vx_1)=0$ but $\qtilde(\vx_1) > 0$, we must find a $\vx_2$ of length $O(\mathrm{poly}(|\vx_1|))$ with $\ptilde(\vx_2) > 0$ but $\qtilde(\vx_2) \leq \qtilde(\vx_1)$.  

To ensure that $\ptilde(\vx_2) > 0$, let us use the structure of $\ptilde$.  For any $j$, we can construct a tautological formula $\phi$ over variables $A_1, \ldots A_j$, as $\phi=(A_1 \lor \neg A_1) \land \cdots \land (A_j \lor \neg A_j)$.  It follows that $\ptilde(\vphi\va) > 0$ for any $\va \in \Bool^j$.  We will take $\vx_2=\vphi\va$ for a particular choice of $j$ and $\va$.

Specifically, we choose them to ensure that $\qtilde(\vx_2) \leq \qtilde(\vx_1)$.  Since $\qtilde$ is ELNCP, it is normalizable and hence has a finite $Z$.  Thus, $\sum_{\va \in \Bool^j} \qtilde(\vphi\va) \leq Z$.  So there must exist some $\va \in \Bool^j$ such that $\qtilde(\vphi\va) \leq \nicefrac{Z}{2^j}$.  We choose that $\va$, after choosing $j$ large enough such that $\nicefrac{Z}{2^j} \leq \qtilde(\vx_1)$. Then $\qtilde(\vx_2) = \qtilde(\vphi\va) \leq \nicefrac{Z}{2^j} \leq \qtilde(\vx_1)$.

To achieve the last claim of the theorem, we must also ensure that $|\vx_2| \in O(\mathrm{poly}(|\vx_1|))$.  Observe that $\qtilde(\vx_1)$ can be computed in polytime (with access to compact parameters), by \cref{thm:closure}.  But this means that the representation of $\qtilde(\vx_1) > 0$ as a rational number must have $\leq g(|\vx_1|)$ bits for some polynomial $g$.  Then $\qtilde(\vx_1) \geq 2^{-g(|\vx_1|))}$, and it suffices to choose $j = \lceil g(|\vx_1|) + \log_2 Z \rceil$ to ensure that $\nicefrac{Z}{2^j} \leq 2^{-g|\vx_1|} \leq \qtilde(\vx_1)$ as required above.  

But then $j \in O(\mathrm{poly}(|\vx_1|))$.  Also, recall that the encoding function $\mathrm{enc}$ used in the construction of $\ptilde$ is guaranteed to have only polynomial blowup (see the proof of \cref{thm:nopforu}).
Thus, $|\vx_2|=|\vphi|+|\va|=|\mathrm{enc}(\phi)|+j \in O(\mathrm{poly}(j)) \subseteq O(\mathrm{poly}(|\vx_1|))$ as required by the theorem.
\end{proof}

\begin{restatable}{locallemma}{noapprox}
	The first part of \Cref{thm:noapproxfullsupport} (without the modifications (a) and (b)).
	
\label{thm:noapprox}
\end{restatable}
We first prove the first part of \cref{thm:noapproxfullsupport} (which is restated in full below).  In this case we will use a distribution $\ptilde$ that does not have support $V^*$ (so it does not prove modification (b)). 
\begin{proof}
We take $\tilde{p}$ to be the weighted language that was defined in \cref{sec:exact}, which was already shown to be efficiently computable.
  Suppose $(\Mq, \vThetaq, \lambda)$ is a counterexample to \cref{thm:noapprox}.  Choose integer $k \geq 1$ in a manner (dependent only on $\lambda$) to be described at the end of the proof. 

Suppose we would like to answer $\textsc{Sat}$ where $\phi$ is a formula with variables $A_1,\ldots,A_j$.  Define $\phi' = (\neg A_{1} \land \neg A_{2} \land \ldots \land \neg A_j \land \neg A_{j+1} \land \neg A_{j+k}) \lor (A_{1} \land \mathrm{Shift}(\phi))$.  Note that $\phi'$ augments $\phi$ with $k$ additional variables, namely $A_1$ and $A_{j+2, \ldots, j+k}$.  For $k=1$, this is the same construction as in the proof of \cref{thm:blowup}.  Let $n=|\vphi'|$ and note that $n$ is polynomial in the size of $\phi$ (holding $k$ constant).

The strings in $\Xc = \mathrm{support}(\ptilde)$ that begin with $\vphi'$ are precisely the strings of the form $\vphi'\va'$ where $\va'$ is a satisfying assignment of $\phi'$.  This is achieved precisely when $\va'=\zero^{j+k}$ or $\va'=\one \va \vec{b}$ where $\va$ is a satisfying assignment of $\phi$ and $\vec{b} \in \mathbb{B}^{k-1}$.  

By our definition of $\ptilde$, all strings in $\Xc$ that begin with $\vphi'$ have equal weight under $\ptilde$.  Call this weight $w$.\footnote{Specifically, each such string has length $n+j+k$, so $\ptilde$ gives it a weight of $w = (\frac{1}{3})^{n+j+k+1}$.} Clearly $Z(\vphi' \zero) = w$, and $Z(\vphi' \one) = w \cdot 2^{k-1} \cdot (\text{number of satisfying assignments of $\phi$})$.  

Recall that $p(\zero \mid \vphi') = Z(\vphi' \zero) / (Z(\vphi' \zero) + Z(\vphi' \one))$.  Let us abbreviate this quantity by $p$.  It follows from the previous paragraph that if $\phi$ is unsatisfiable, then $p = 1$, but if $\phi$ is satisfiable, then $p \leq  \nicefrac{1}{(1+2^{k-1})}$.  By hypothesis, $p$ is approximated (with error probability $< \nicefrac{1}{3}$) by the possibly random quantity
$(\Mq(\vthetaq_{|\vphi'|}))(\vphi' \zero)$, which we abbreviate by $q$, to within a factor of $\lambda$.  That is, $p \in [q/\lambda, \lambda q]$.  By choosing $k$ large enough\footnote{It suffices to ensure that $1+2^{k-1} > \lambda^2$, so take any $k > 1+ \log_2 (\lambda^2-1)$.}  such that $[q/\lambda, \lambda q]$ cannot contain both $1$ and $\nicefrac{1}{(1+2^{k-1})}$, we can use $q$ to determine whether $p=1$ or $p \leq \nicefrac{1}{(1+2^{k-1})}$.  This allows us to determine $\textsc{Sat}(\phi)$ in polynomial time with error probability $< \nicefrac{1}{3}$, since by hypothesis $q$ is computable in polynomial time with compact parameters.  This shows that $\textsc{Sat} \in \mathrm{BPP/poly} = \mathrm{P/poly}$, implying $\mathrm{NP} \subseteq \mathrm{P/poly}$, contrary to our assumption.  ($\mathrm{BPP/poly}$ is similar to $\mathrm{P/poly}$ but allows $\Mq$ to be a bounded-error probabilistic Turing machine.)
\end{proof}
\noapproxfullsupport*
\begin{proof} 
It remains to show that the statement remains true with modification (a) and with modification (b).  For (a), the proof of \cref{thm:noapprox} suffices, since it reduces \textsc{Sat} to approximate local probability queries of the stated form.  That is, the true local probabilities $p(x \mid \vxhat)$ that can be computed with finite summations, thanks to the structure of our example language $\ptilde$, which guarantees that the prefix $\vxhat$ can only continue with suffixes of a fixed length that is easily determined from $\vxhat$.

For modification (b), again let $V = \Bool = \{\zero,\one\}$.  Choose some $\epsilon > 0$ (any choice will do), and let
\begin{align}
\ptilde_1(\vx) &= 
   \begin{cases}
      (\tfrac{1}{3})^{|\vx+1|} & \text{if } \vx = \vphi\va \text{ where } \vphi=\mathrm{enc}(\phi) \\ & \qquad \text{ and } \va \text{ satisfies }\phi \\
      0 & \text{otherwise}
   \end{cases} \nonumber \\
\ptilde_2(\vx) &= (\tfrac{1}{9})^{|\vx+1|} > 0 \nonumber\\
\ptilde(\vx) &= \ptilde_1(\vx) + \epsilon \cdot \ptilde_2(\vx) \nonumber
\end{align}
We use $Z_1$, $Z_2$, and $Z$ respectively to denote normalizing constants of these three weighted languages.  Note that $\ptilde_1$ is the weighted language that was previously used in the proofs of \cref{thm:blowup,thm:noapprox}.  Our new $\ptilde$ is intended to be very similar while satisfying the additional condition (b).  It is easy to show that $\ptilde$ is efficiently computable, much as we showed for $\ptilde_1$ in \cref{thm:blowup}.  Also, $\ptilde$ is normalizable, since $Z = Z_1+\epsilon\cdot Z_2$, where $Z_1 \leq (\frac{1}{3}) / ( 1 - \frac{2}{3} ) = 1$ and $Z_2 = (\frac{1}{9}) / (1 - \frac{2}{9}) = \frac{1}{7}$ are both finite.

The proof proceeds as in \cref{thm:noapprox}, with $\phi'$ constructed from $\phi$ as before.  Recall that $\phi$ has $j$ variables, $\phi'$ has $j+k$ variables, and $|\vphi'|=n$.  We may assume WLOG that the encoding function $\mathrm{enc}$ is such that an encoded formula always has at least as many bits as the number of variables in the formula, so $n \geq j+k$.  

Notice that $Z_1(\vphi')$ sums over the satisfying assignments of $\phi'$, and there may be as few as one of these (if $\phi$ is unsatisfiable).  By contrast, $Z_2(\vphi')$ sums over an infinite number of continuations with positive probability.  The faster decay rate of $\frac{1}{9}$ in $\ptilde_2$ was chosen to keep $Z_2(\vphi')$ small relative to $Z_1(\vphi')$ despite this.  Specifically,
\begin{align*}
Z_1(\vphi' \zero) &= (\tfrac{1}{3})^{n+j+k+1} \\
Z_1(\vphi' \one) &= (\tfrac{1}{3})^{n+j+k+1} \cdot 2^{k-1} \\ & \qquad \cdot (\text{\# of satisfying assignments of }\phi) \\
Z_2(\vphi' \zero) &= (\tfrac{1}{9})^n \cdot \tfrac{1}{9} \cdot (\tfrac{1}{9} / (1 - \tfrac{2}{9})) \\
                 & = \tfrac{1}{7} \cdot (\tfrac{1}{3})^{2(n+1)} \\
                 &< \tfrac{1}{7} \cdot Z_1(\vphi' \zero) \\ & \text{ (because $2(n+1) > n+j+k+1$)} \\
Z_2(\vphi' \one) &= Z_2(\vphi' \zero)
\end{align*}
As in the proof of \cref{thm:noapprox}, we will show that $p(\zero \mid \vphi')$ is much larger when $\phi$ is unsatisfiable.  Recall that $Z(\vxhat) = Z_1(\vxhat) + \epsilon \cdot Z_2(\vxhat)$.  When $\phi$ has zero satisfying assignments,
\begin{align*}
p(\zero \mid \vphi') &= \frac{Z(\vphi' \zero)}{Z(\vphi' \zero) + Z(\vphi' \one)} 
                    \\ & = \frac{Z(\vphi' \zero)}{Z_1(\vphi' \zero) + \epsilon\cdot Z_2(\vphi' \zero) + \epsilon\cdot Z_2(\vphi' \one)}
                    \\ & > \frac{Z(\vphi' \zero)}{Z_1(\vphi' \zero) + 2\cdot \tfrac{\epsilon}{7} \cdot Z_1(\vphi' \zero)} 
\end{align*}
whereas if $\phi$ has at least one satisfying assignment, then
\begin{align*}
p(\zero \mid \vphi') &= \frac{Z(\vphi' \zero)}{Z(\vphi' \zero) + Z(\vphi' \one)} 
                   \\ &  < \frac{Z(\vphi' \zero)}{Z_1(\vphi' \zero) + Z_1(\vphi' \one)} 
                \\ &     \leq \frac{Z(\vphi' \zero)}{Z_1(\vphi' \zero) + 2^{k-1} Z_1(\vphi' \zero)} 
\end{align*}
This rewrites both probabilities in terms of $Z_{\cdot}(\vphi'0)$ quantities, which do not depend on the number of satisfying assignments.  So now we can see that the first probability is at least $(1+2^{k-1}) \;/\; (1+\frac{2\epsilon}{7})$ times as large as the second probability.
Choose $k$ large enough\footnote{It suffices to ensure that $(1+2^{k-1}) / (1+\frac{2\epsilon}{7}) > \lambda^2$, so take any $k > 1+ \log_2 (\lambda^2 \cdot (1+\frac{2\epsilon}{7}) - 1)$.} such that $[q/\lambda, \lambda q]$ cannot contain both probabilities, and complete the proof as in \cref{thm:noapprox}.  
\end{proof}

\setcounter{theorem}{4}
\begin{theorem}
	The set $\{\ptilde: \ptilde \text{ is normalizable}, \ptilde \in \mathrm{EC}, \ptilde \not\in\mathrm{ELN} \}$ is not empty.
	\label{thm:in-ec-not-in-eln}
\end{theorem}

\cref{thm:in-ec-not-in-eln} states that some normalizable EC distributions cannot be expressed as ELN distributions.  
The proof is based on the 
undecidability
of the halting problem, rather than the assumed inefficiency of the Boolean satisfiability problem.  Thus, unlike \cref{thm:blowup}, it does \emph{not} rely on the assumption that $\npppoly$, or even on the weaker assumption that $\mathrm{P}\neq\mathrm{NP}$.

\begin{proof}
	Given any unweighted language $L \subseteq \Bool^*$, we can define a normalizable weighted language $\ptilde$ with support $L$ by $\ptilde(\vx) = \nicefrac{1}{3}^{|\vx|+1}$ for $\vx \in L$ and $\ptilde(\vx) = 0$ otherwise.  Moreover, if $L \in \mathrm{P}$, then $\ptilde \in \mathrm{EC}$.
	
	For our purposes, we take $L$ to consist of all strings of the form $\vx^{(1)} \vx^{(2)}$, for which there exists a deterministic Turing machine $M$ such that $\vx^{(1)} = \mathrm{enc}(M)$ (where $\mathrm{enc}$ is a prefix-free encoding function) and $\vx^{(2)}$ encodes an accepting execution path of $M$ on an empty input.  (Such a path may be represented as a sequence of transitions of $M$ that begins with an initial state and ends at an accepting state.)  Note that any deterministic TM $\vx^{(1)}$ can be paired with at most one accepting execution path $\vx^{(2)}$, and cannot be paired with any $\vx^{(2)}$ if it does not halt.
	
	Clearly $L\in\mathrm{P}$: given $\vx \in \Bool^*$, we can decide whether $\vx \in L$ by first checking if $\vx$ can be expressed as a concatenation of strings $\vx^{(1)}$ and $\vx^{(2)}$ of the required form. Then we build $M$ from $\vx^{(1)}$ and simulate it to check the transitions in $\vx^{(2)}$ on $M$ step-by-step.  This can be done in $O(\mathrm{poly}(|\vx|))$ total time.  We conclude that the $\ptilde$ derived from $L$ is EC.
	
	Now, $Z(\vx^{(1)}) > 0$ iff $M$ halts on the empty input.
	But this undecidable problem could be decided if there were an ELN weighted language that had support $L$, since then $Z(\vx^{(1)})\,/\,Z$ could be found as a product of local conditional probabilities, $\prod_{t=1}^{|\vx^{(1)}|} p(x^{(1)}_t \mid \vx^{(1)}_{<t})$, that could each be computed by a Turing machine.  Therefore $\ptilde$ is not ELN.
\end{proof}
	
We have shown above that a certain unweighted language $L$ is not the support of any ELN distribution.  We conjecture that it is also not the support of any ELNCP distribution;\footnote{We have not attempted to prove this.  Our loose intuition is that the compact parameters of an ELNCP language may help it to memorize some small part of $L$, but the halting problem would still be undecidable when restricted to the rest of $L$ \cite{10.2307/27588653}.} a proof of this would strengthen \cref{thm:in-ec-not-in-eln} to become an unconditional version of \cref{thm:blowup}.  However, ELNCP weighted languages do have more power than ELN weighted languages, as we now show.

\begin{theorem}
	The set $\{\ptilde: \ptilde \text{ is normalizable}, \ptilde \in \mathrm{EC}, \ptilde \in \mathrm{ELNCP}, \ptilde \not\in\mathrm{ELN} \}$ is not empty.
	\label{thm:in-ec-in-elncp-not-in-eln}
\end{theorem}

\Cref{thm:in-ec-in-elncp-not-in-eln} justifies why this region is drawn as non-empty in \cref{fig:modelsupportzoo}.  Again, it does 
not rely on the assumption $\npppoly$ or $\mathrm{P}\neq \mathrm{NP}$.
Note that \cref{thm:in-ec-not-in-eln} can be regarded as a corollary of \cref{thm:in-ec-in-elncp-not-in-eln}. 

\begin{proof}
    The weighted language $\ptilde$ constructed in \cref{thm:in-ec-not-in-eln} is not necessarily ELNCP.  To fix this, we modify the construction to obtain a weighted language $\ptilde'$ with \emph{sparse} support $L'$.  We will again be able to show that $\ptilde'$ is EC and not ELN.  To show that $\ptilde'$ is also ELNCP, we will rely on the sparsity of $L'$, meaning that $\mathrm{prefixes}(L') \defeq \{\vxhat': (\exists \vx' \in L')\;\vxhat' \prefixof \vx'\}$ contains at most $O(\mathrm{poly}(n))$ strings $\vxhat'$ of length $\leq n+1$.  Thus, we can use $\vThetaq_n$ to store all of those strings $\vxhat'$ in polynomial space, along with their $Z(\vxhat')$ values.\footnote{More precisely, the first $b$ bits of $Z(\vxhat') \leq 1$ may be stored in $\vThetaq_{n+b}$, when ELNCP is defined as explained in our ``Remark on irrationality'' above.}  Notice that all strings $\vxhat' \notin \mathrm{prefixes}(L')$ have $Z(\vxhat')=0$, so they need not be stored.  Now for any $\vxhat'$ of length $\leq n$, a Turing machine that consults $\vthetaqn$ can compute $q(x \mid \vxhat') = Z_{\ptilde'}(\vxhat' x)\,/\,Z_{\ptilde'}(\vxhat')$ in time $O(\mathrm{poly}(n))$ as desired, establishing that $\ptilde'$ is ELNCP.

	We may define $\ptilde'$ as follows.  Let $\mathrm{sparsify}(\vx)$ be a version of $\vx$ with many extra {\zero} symbols inserted: specifically, it inserts $2^t$ copies of {\zero} immediately before the $t$\textsuperscript{th} bit of $\vx$, for all $1 \leq t \leq |\vx|$.  We construct $\ptilde'$ so that $\ptilde'(\mathrm{sparsify}(\vx)) = \ptilde(\vx)$. Specifically, let $L' \defeq \mathrm{sparsify}(L)$.   The inverse function $\mathrm{sparsify}^{-1}(\vx')$ is defined on exactly $\vx' \in L'$, and is unique when defined. For all $\vx' \in \Bool^*$, let $\ptilde'(\vx') \defeq  \ptilde(\mathrm{sparsify}^{-1}(\vx'))$ if  $\mathrm{sparsify}^{-1}(\vx')$ is defined, and $\ptilde'(\vx') \defeq 0$ otherwise.  This can be computed in polytime, so $\ptilde'$ is EC.  Also, its support $L'$ is sparse as claimed, so $\ptilde'$ is ELNCP.

	Finally, we claim $\ptilde'$ is not ELN. A given deterministic Turing machine $M$ halts on the empty input iff $\mathrm{enc}(M) \in \mathrm{prefixes}(L)$ iff $\mathrm{sparsify}(\mathrm{enc}(M)) \in \mathrm{prefixes}(L')$ iff $Z'(\mathrm{sparsify}(\mathrm{enc}(M))) > 0$.
	But as in the proof of \cref{thm:in-ec-not-in-eln}, this would be decidable if $\ptilde'$ were ELN as defined in \cref{sec:local-normalization}, since then we would have a Turing machine to compute the local conditional probabilities $p'(\hat{x}_t \mid \vxhat_{<t})$ for $\vxhat = \mathrm{sparsify}(\mathrm{enc}(M))$.
\end{proof}

\evalmarginalized*
\begin{proof}
	We will construct $p$ such that $\textrm{support}(p)$ is the $\mathrm{NP}$-complete language \textsc{Sat} of all satisfiable boolean formulas.  The idea is to construct an ELN distribution $r$ that can autoregressively generate any assignment $\va$ followed by any formula $\phi$ that is satisfied by $\va$.  Thus, if we delete the $\va$ prefixes, the support consists of exactly the satisfiable formulas $\phi$ (or more precisely, their encodings $\vphi$).  
	
	To be more precise, we will have $\mathrm{support}(r)$ be the language $L = \{ \va\sep\vphi \mid \va \in \Bool^* \text{ and } \phi \text{ is a formula satisfied by }\va \}$.  This is defined similarly to the support language $L$ in \cref{sec:exact}, but with the order of $\vphi$ and $\va$ crucially swapped: $r$ will now generate the ``solution'' $\va$ \emph{before} the ``problem'' $\vphi$.  The alphabet $V$ of this language contains at least the symbols $\{\zero,\one,\sep\}$, where \sep is a separator symbol, and any other symbols needed to encode $\phi$ as $\vphi$.  The marginalization operator $\mu$ maps $\va\sep\vphi$ to $\vphi$.
	
	Let $j=|\va|$.  As in \cref{sec:exact}, we will require $\phi$ to use all of the variables $A_1,\ldots,A_j$ (and only those variables), implying that $|\vphi| \geq j$.  This ensures that marginalizing over the $j+1$ latent symbols is only light marginalization since $j+1+|\vphi| \in O(\mathrm{poly}(|\vphi|))$.  For convenience, we will also require $\phi$ to be a CNF formula.  These requirements shrink $\textrm{support}(p)$ but do not affect its $\mathrm{NP}$-completeness.
	
	The remaining challenge is to construct an autoregressive distribution $r$ whose support is $L$.  We can think of this distribution as describing an efficient procedure for randomly generating a string from left to right so that the procedure generates the $t$\textsuperscript{th} symbol in time $O(\mathrm{poly}(t))$, terminates with probability 1,\footnote{\label{fn:termination}Phase 1 almost surely terminates after a finite number of bits.  Phase 2 almost surely terminates after a finite number of clauses, and each clause almost surely terminates after a finite number of literals. ``Almost surely'' means ``with probability 1.''} has positive probability of producing any string in $L$, and has zero probability of producing any string not in $L$.  Below we give such a procedure.\footnote{Our presentation here makes use of an infinite alphabet that includes symbols such as $A_i$ and $\neg A_i$ for all $i \in \mathbb{N}_{>0}$, as well as symbols such as $\zero, \one, \land, \lor$.  We implicitly invoke some prefix-free encoding scheme to translate each symbol into a fixed string over the finite alphabet $V$.  
	}
	
	\begin{enumerate}
        \item First, the procedure generates $\va\sep$ as a sequence of  random symbols from $\{\zero,\one,\sep\}$, making a uniform draw at each step.  It stops immediately after generating $\sep$ for the first time.  The string generated before $\sep$ is called $\va$ and we let $j=|\va|$.  For example, $\va=\zero\one\zero$ and $j=3$.
    
    \item Second, the procedure must generate the encoding $\vphi$ of a random CNF formula $\phi$ that is satisfied by $\va$, such as $(A_2 \lor \neg A_3 \lor \neg A_2 \lor A_2) \land (\neg A_1)$ in our example.  This involves generating a random sequence of 0 or more satisfied clauses connected by $\land$.  At each step, the procedure decides whether to generate a new clause or end the formula.  The probability of generating a new clause is ordinarily $\nicefrac{1}{2}$.  However, this probability is $1$ if the previous clauses do not yet mention all the variables $A_1,\ldots,A_j$.
    
    How does it generate each satisfied clause?  This involves generating a sequence of literals connected by $\lor$, at least one of which must be true.  
    At each step of this subroutine, it uniformly chooses an integer $i \in [1,j]$, and then flips a fair coin to decide whether to add the literal $A_i$ or $\neg A_i$ to the current clause. If the clause is now satisfied by $\va$ (\emph{i.e.}, at least one of the literals is true), it then flips another fair coin to decide whether to end the clause.
    \end{enumerate}
    
    $r$ is ELN because there exists a Turing machine that computes from input $\vxhat x$\Dash in time $O(\mathrm{poly}(|\vxhat|))$\Dash the probability that the next symbol generated after the prefix $\vxhat$ would be $x$, under the above procedure.  
    As discussed in \cref{fn:consistency}, that probability equals $r(x \mid \vxhat)$\Dash which is what our Turing machine is required to return\Dash because the above procedure almost surely terminates (\cref{fn:termination}), ensuring that $r$ is a consistent probability distribution over $V^*$ (that is, $\sum_{\vx \in V^*} r(\vx) = 1$). 
\end{proof}

\nppolyifflmeccp*
\begin{proof}
    (b) implies (c) since any ELNCP distribution is an ECCP weighted language (\cref{thm:closure}). (c) implies (a) by \cref{thm:marginalized-to-np-poly} below. 
    Finally, (a) implies (b) by \cref{thm:eccp-to-marginalized} below. 
\end{proof}

\begin{restatable}{locallemma}{marginalizedtonppoly}
For any ECCP weighted language $\rtilde$, if $\ptilde$ is a light marginalization of $\rtilde$, then $\mathrm{support}(\ptilde) \in \mathrm{NP/poly}$.
\label{thm:marginalized-to-np-poly}
\end{restatable}

Notice that this lemma concerns the class $\mathrm{NP/poly}$, not $\mathrm{P/poly}$ (see \cref{sec:p-poly}).  The proof is straightforward.

\begin{proof}
Suppose $\rtilde$ is ECCP via $(\Mtilder,\vThetatr)$, and $\mu$ is the  marginalization operator such that $\ptilde(\vx) = \sum_{\vz: \mu(\vz) = \vx} \rtilde(\vz)$.  By the light marginalization assumption, there is a polynomial $f$ such that $|\vz| \leq f(|\mu(\vz)|)$.

To prove $\mathrm{support}(\ptilde) \in \mathrm{NP/poly}$, we must show that there exists $(M,\vTheta)$
such that for all $n \geq 0$, a \emph{nondeterministic} Turing machine $M_n$ can be constructed as $M(\vtheta_n)$ in time $O(\mathrm{poly}(n))$, which can in turn decide in time $O(\mathrm{poly}(n))$ whether $\ptilde(\vx) > 0$ for any $\vx$ with $|\vx|=n$. 

Deciding $\ptilde(\vx) > 0$ means deciding whether $(\exists \vz \in V^*)\; \mu(\vz) = \vx \text{ and } \rtilde(\vz) > 0$.  But if $|\vx|=n$, the first condition $\mu(\vz)=\vx$ implies $|\vz| \leq f(|\mu(\vz)|) = f(|\vx|)=f(n)$.  Thus, we need $M_n$ to nondeterministically check only the $\vz$ of length up to $f(n)$ to see whether $\mu(\vz)=\vx$ and $\rtilde(\vz)>0$.  

How can $M_n$ check a string $\vz$ of length $m$? It can decide the first condition $\mu(\vz)=\vx$ in time $O(\mathrm{poly}(m))$, since the marginalization operator $\mu$ is a polytime function.  To decide the second condition $\rtilde(\vz)>0$, it must construct the (deterministic) Turing machine $\Mtilder(\vthetatr_m)$ and then apply it to $\vz$ to obtain $\rtilde(\vz)$: since $\rtilde$ is ECCP, both steps take time $O(\mathrm{poly}(m)) = O(\mathrm{poly}(f(n))) \subseteq O(\mathrm{poly}(n))$ as required.

However, this means that $M_n = M(\vtheta_n)$ must have access to the parameter vectors $\vthetatr_m$ for all $m \leq f(n)$.  We therefore make $\vtheta_n$ include this collection of parameter vectors.  Each $|\vthetatr_m| \in O(\mathrm{poly}(m)) \subseteq O(\mathrm{poly}(n))$ since $\rtilde$ is ECCP. So $|\vtheta_n| \in O(\mathrm{poly}(n))$ as required.
\end{proof}

\begin{restatable}{locallemma}{eccptomarginalized}
For any $L \in \mathrm{NP/poly}$, there exists a light marginalization $p$ of an ELNCP distribution, such that $\mathrm{support}(p) = L$.
\label{thm:eccp-to-marginalized}
\end{restatable}

\Cref{thm:eccp-to-marginalized} resembles \cref{thm:evalmarginalized}, but it constructs distributions for \emph{all} $L \in \mathrm{NP/poly}$, not just for \emph{one particular} $L \in \mathrm{NPC}$.  The proof is similar but more complicated.  In both cases, the goal is to demonstrate how an ELNCP distribution $r$ can define a left-to-right stochastic string generation process such that the suffix of the generated string must be in $L$ and can be any element of $L$.

Our string generation process in this case is inspired by rejection sampling, a widely used method for sampling from an energy-based model with support $L$. The standard scheme is to first sample a string $\vx$ from a tractable distribution $q$ such that $\mathrm{support}(q) \supseteq L$, then accept the sample with an appropriate probability, which is 0 if $\vx \notin L$.  The process is repeated until a sample is finally accepted.  There is no guarantee that this standard scheme will terminate in polynomial time, however.  
Fortunately, in our setting, we are not trying to match our sampling distribution $p$ to a given energy-based model, but simply match its support to a given language $L$.  We make use of the polysize parameter vectors of ELNCP languages to store certain `fallback strings' that are guaranteed to be in the desired language $L$. Wherever ordinary rejection sampling would reject a string and try generating another, we switch to generating a stored fallback string of an appropriate length.  This scheme places all of the rejected probability mass on the small set of fallback strings (in contrast to rejection sampling, which in effect throws away this mass and renormalizes).  %
The advantage is that it does not iterate indefinitely. 
At a high level, $r$ is a distribution over strings $\vz$ that record traces of this generative story we describe above. 

\begin{proof}
	 WLOG we assume $L$ uses the alphabet $V = \{\zero, \one, \sep\}$.
	 In the case where $L$ is finite, the result is trivial.  We simply define $r(\vx) = \nicefrac{1}{|L|}$ for $\vx \in L$  and $r(\vx) = 0$ otherwise. We then take $p=r$ (a trivial marginalization).  It is easy to show that $r$ is ELN, and therefore ELNCP as desired, by constructing an appropriate Turing machine that maps $\vxhat x$ to $r(x \mid \vxhat)$ in time $O(|\vxhat x|)$, for any $\vxhat$ that is a prefix of some string in $L$ and any $x \in V \cup \{\eos\}$.  The finite state table of the Turing machine includes states that correspond to all possible strings $\vxhat x$, with transitions arranged in a trie.  It reads the input string $\vxhat x$ from left to right to reach the state corresponding to $\vxhat x$.  If it detects the end of the input while in that state, it writes $r(x \mid \vxhat)$ on the output tape.

	 Now we consider the case where $L$ is infinite.  For each $j \in \mathbb{N}_{\geq 0}$, let the `fallback string' $\vx^{(j)}$ be some string in $L$ of length $\geq j$.  For definiteness, let us take it to be the shortest such string, breaking ties lexicographically.  At least one such string does exist because $L$ is infinite, so $\vx^{(j)}$ is well-defined.
	 
	 Also, since $L \in \mathrm{NP/poly}$ (\cref{sec:p-poly}), let $(M,\vTheta)$ be an ordered pair and $f$ be a polynomial such that $M_j = M(\theta_j)$  nondeterministically accepts $\va$ within $\leq f(j)$ steps iff $\va \in L$.

	As in the proof of \cref{thm:evalmarginalized}, we now describe a procedure for randomly generating a string $\vz$ from left to right. $\vz$ will have the form $\va\sep\vb\sep c \vd$, where $\vd \in L$ and the latent substring $\va\sep\vb\sep c$ will be removed by the marginalization operator $\mu$.
	\begin{enumerate}
	\item First we generate a random string $\va \in \Bool^*$ followed by $\sep$, just as in the proof of \cref{thm:evalmarginalized}. Again let $j=|\va|$.  
	\item Next, we must consider whether $\va \in L$.  We generate a random computation path $\vb$ of $M_j$ on input $\va$ until it either accepts (in which case we then generate $\sep \one$ to record acceptance of $\va$) or has run for $f(j)$ steps without accepting (in which case we then generate $\sep \zero$ to record rejection).  
	\item In the former case ($c=\one$) we finish by deterministically generating $\vd \defeq \va \in L$.  In the latter case ($c=\zero$), $\va \notin L$, so we fall back and finish by deterministically generating $\vd \defeq \vx^{(j)} \in L$.  
    \end{enumerate}

    Let $r(\vz)$ be the probability that the above procedure generates $\vz$. 
    $\mathrm{support}(r)$ is then the set of strings that can be generated by the above procedure.  The marginalized language $\mu(\mathrm{support}(r))$ keeps just the $\vd$ parts of those strings.  It consists of all strings $\va$ that are accepted by at least one path $\vb$ of $M_{|\va|}$ (which are exactly the strings in $L$) together with the fallback strings (which form a subset of $L$).  Thus, $\mu(\mathrm{support}(r))=L$ as desired.

    We wish to show that $r$ is ELNCP.  In other words, some Turing machine $\Mq$ efficiently locally normalizes $r$ with compact parameters $\vThetaq$, as defined in \cref{sec:local-normalization}.  The parameters will be used to store information about the infinite set of fallback strings.  
    
    In particular, for each $n$, $\vthetaqn$ must have enough information to construct a Turing machine $q_n = \Mq(\vthetaqn)$ such that $q_n(\vzhat z)$ returns $r(z \mid \vzhat)$ for all $z \in V \cup \{\eos\}$ and all $\vzhat$ with $|\vzhat| \leq n$ and $Z(\vzhat) > 0$.  Here $Z(\vzhat) > 0$ means that $\vzhat$ is a prefix of a string $\vz = \va\sep\vb\sep c \vd$ that could be generated by the above procedure.  The computation $q_n(\vzhat z)$ proceeds by simulating the sequence of choices in the above procedure that would be required to generate $\vzhat$, and then returning the probability that the procedure would generate symbol $z$ next.  That probability equals $r(z \mid \vzhat)$ as desired because the above procedure almost surely terminates (as explained at the end of the proof of \cref{thm:evalmarginalized}).  
    
    In general, the computation $q_n(\vzhat z)$ may have to construct $M_j = M(\theta_j)$ and simulate it on $\va$ (for $j=|\va|$) if $z$ falls in the $\vb\sep c$ portion of $\vzhat$, and it may have to look up a character of the fallback string $\vx^{(j)}\eos$ if $z$ falls in the $\vd$ portion of $\vzhat$ or terminates that portion with $z=\eos$.  Fortunately $j < n$, and fortunately if the computation looks up the $t$\textsuperscript{th} character of $\vx^{(j)}\eos$ then $t < n$.  Thus, constructing and simulating $M_j$ can be done in time $O(\mathrm{poly}(j)) \subseteq O(\mathrm{poly}(n))$, and looking up the $t$\textsuperscript{th} character of $\vx^{(j)}\eos$ can be achieved with access to the first $n$ characters of each of $\vx^{(1)}, \ldots, \vx^{(n)}$, which can be stored by $\vthetaqn$ in space $O(n^2)$. It follows that $\Mq$ can construct and apply $q_n$ in polynomial time with access to compact parameters $\vThetaq$, so $r$ is ELNCP.

\end{proof}

\section{Implementation details of REBMs}
\label{sec:rlm-details}
\subsection{Modeling finite subsets of infinite languages}
\label{sec:finitegrowth}
The experiments of this paper are conducted on datasets where we only observe strings that are finitely long.  Given a possibly infinite language $\Xc$, we use the notation $\Xc_{\leq T} = \{ \vx \mid \vx \in \Xc, |\vx| \leq T \}$ for the subset of strings that are most $T$ symbols long. Specific values of $T$ for datasets used in our experiments are listed in \cref{sec:datasets}.

\subsection[whatever2]{Design of base models $p_0$}
\label{sec:baseline-design}
$p_0$ can be any distribution over $\Xc_{\leq T}$\footnote{Note that since $p_0$ does not have support over $\Xc$, it has to assign $p(\eos \mid \vx_{1\ldots T})=1$, which is generally not an issue.} provided that we can sample from it, and evaluate $p_0(\vx), \forall \vx \in \Xc_{\leq T}$, both in $O(\mathrm{poly}(|\vx|))$. In this work, we experiment with two designs of $p_0$: GRU- and Transformer-based locally normalized language models.
GRU-based models are used in WikiText and Yelp experiments. 
The GRU-based $p_0$'s are parametrized with $2$-layer GRUs with $500$ hidden units, and word embeddings of dimension size $500$.

As for Transformer-based $p_0$'s, we make use of Grover models \cite{zellers2019defending},
which effectively are GPT-2 models trained on the aforementioned \textsc{RealNews} dataset. In this work, we experiment with the `base' variant of public available weights, which are $12$-layered Transformers, with $12$ heads, and $768$ hidden units. %

\subsection[whatever3]{Design of discriminators $\g$}
\label{sec:discriminator-design}
We formulate $\g(\vx)$ as a summation of scores at positions $1 \ldots |\vx|$, passed through an activation function $f$:
\begin{align}
    \g(\vx) =  f\left( \sum_{i=1}^{|\vx|} g_t(\vx; \vtheta) \right).
\end{align}
To verify whether lower-bounding $\g$ would help with learning, as we discuss in \cref{sec:ebm}, we experiment with two variants of $f$:
\begin{itemize}
        \item {\bf tanh}: $f(x) = 2 \cdot \tanh(x)$
        \item {\bf softplus}: $f(x) = - \log (1 + \exp(x + s))$
    \end{itemize}
The former one is bounded between $(-2, 2)$, while the second one has range $(-\infty, 0)$. The offset term $s$ in the softplus activation function determines initial values of $\Z$. In this paper we set $s = 20$.

The design of $g_{t}(\vx; \vtheta)$ follows their base model counterparts: we use Bi-GRU discriminators for GRU base models; and bi-directional Transformer discriminators for Transformer ones. For GRUs $g_t(\vx; \vtheta) = \vh_t \cdot x_t$, For Transformers $g_t(\vx; \vtheta) = \sum \vh_t$  where $\vh_t$ are the hidden states at time step $t$. In both cases, the discriminators have access to information of the whole sequence $\vx$ at any timestep: the Bi-GRU discriminators achieve this through the bi-directional RNNs, and the Transformers through the attention mechanism without directional masking.
\subsection{Training procedure}
\label{sec:training-procedure}
As we note in \cref{sec:ebm}, MLE-based training methods are generally not feasible for globally normalized models.
We therefore opt to train our model using the ranking variant of noise contrastive estimation (NCE) \citep{Ma2018NoiseCE},
which does not require samples from $p_0$ and has a simple form for residual LMs.
Using $p_0$ as a \emph{noise distribution}, NCE training requires minimizing the following single-sequence loss, in expectation over the true distribution $p$:
\begin{align}
    \mathcal{L}_{\textsc{nce}}(\vtheta, \vx, p_0, K) = - \log \frac{\frac{\tilde{p}_{\vtheta}}{p_0}(\vx)}{ \sum_{k=0}^{K} \frac{\tilde{p}_{\vtheta}}{p_0}(\vx^{(k)}) },
    \label{eq:nce-loss}
\end{align}
where $\vx^{(0)} \triangleq \vx$, $\frac{\tilde{p}_{\vtheta}}{p_0}(\vx) \triangleq \frac{\tilde{p}_{\vtheta}(\vx)}{p_0(\vx)}$, and $\vx^{(1)} \ldots \vx^{(K)} \sim p_0$. 
Since $\tilde{p}_{\vtheta}(\vx) = p_0(\vx) \cdot \exp \g(\vx)$, we have $\frac{\tilde{p}_{\vtheta}}{p_0}(\vx) = \exp \g(\vx)$.  The NCE minimization objective \eqref{eq:nce-loss} now reduces to the simple form
\begin{align}
   \MoveEqLeft[1]  \mathcal{L}_{\textsc{nce}}(\vtheta, \vx, p_0, K) &  \nonumber \\ %
     & = - \g(\vx)  \nonumber \\ & + \log (\exp \g(\vx) + \sum_{k=1}^K \exp g_{\vtheta} (\vx^{(k)})).
    \label{eq:rlm-nce-loss}
\end{align}

Notice that minimizing the expected loss with stochastic gradient descent methods 
$\mathcal{L}_{\textsc{nce}}$ defined in \cref{eq:rlm-nce-loss} requires only evaluating sequence probabilities under $\g$, and tuning its parameters, but not the base model $p_0$. 
We only need to generate the noise samples $\{ \vx^{(k)} \sim q \mid k \in [K] \}$ from $p_0$. This way we do not need to backpropagate through parameters of the base model $p_0$, which can speed up training considerably when $p_0$ is backed by a huge network. In fact, the training of $\g$ can be completely agnostic to the design of $p_0$, allowing for the application of finetuning any locally normalized $p_0$.

Given the same discriminator $\g$, the difference of $\mathrm{KL}$-divergence between the true model $p$ and residual language models $\tilde{p}'_{\vtheta}(\vx) = p'_0(\vx) \cdot \exp \g(\vx)$, and the $\mathrm{KL}$-divergence between the true model  and $\tilde{p}''_{\vtheta}(\vx) = p''_0(\vx)\cdot \exp \g(\vx)$, defined with base models $p'_0$ and $p''_0$ respectively, can be written as
\begin{align} \MoveEqLeft[4]
    \mathrm{KL}[p|| {p}'_{\vtheta} ] - \mathrm{KL}[p || {p}''_{\vtheta} ] & \nonumber \\ &= \mathrm{KL}[p || p'_0] -  \mathrm{KL}[p|| p''_0 ] + \log \frac{Z'}{Z''},
    \label{eq:kl-relation}
\end{align}
where $Z' = \E[\vx \sim p'_{0}]{\exp \g(\vx)}$, and $Z''$ is similarly defined with $p''_0$.
As a direct result of \cref{eq:kl-relation}, we can see that finding $p''_0$ where $\mathrm{KL}[p||p''_0] < \mathrm{KL}[p||p'_0]$ implies improvement in $\mathrm{KL}[p||p''_{\vtheta}]$ over $\mathrm{KL}[p||p'_{\vtheta}]$, under mild conditions:
\begin{restatable}{localtheorem}{bound}
If $\exists k > 0$ such that
$\frac{\E[\vx \sim p'_0]{\exp \g(\vx)}}{ \E[\vx \sim p''_0]{\exp \g(\vx) }  } > \exp(-k)$ and
${\mathrm{KL}} [p || p'_{0}] - {\mathrm{KL}} [p || p''_{0}] > k$ then ${ \mathrm{KL} [p || p'_{\vtheta}] > \mathrm{KL} [p||p''_{\vtheta}]} $.
\label{propo:bound}
\end{restatable}
\begin{proof}
\begin{align}
& \mathrm{KL}[p || p'_{\vtheta}] - \mathrm{KL}[p || p''_{\vtheta}]  \nonumber \\&= \E[\vx \sim p]{\log p''_{\vtheta}(\vx) - \log p'_{\vtheta}(\vx) } \nonumber \\
&= \E[\vx \sim p]{\log \frac{p''_0(\vx) \exp \g(\vx) }{\sum_{\vx' \in \Xc_{\leq T}} p''_0(\vx) \exp \g(\vx) } \nonumber \\ & - \log \frac{ p'_0(\vx) \exp \g(\vx) }{\sum_{\vx' \in \Xc_{\leq T}} p'_0(\vx) \exp \g(\vx) }  } \nonumber \\
&= \E[\vx \sim p]{\log \frac{ p''_0(\vx) \exp \g(\vx) }{\E[\vx' \sim p''_0]{ \exp \g(\vx)} } \nonumber \\ & - \log \frac{ p'_0(\vx) \exp \g(\vx) }{\E[\vx' \sim p'_0]{ \exp \g(\vx)} }} \nonumber \\
&= \E[\vx \sim p]{\log p''_0(\vx) - \log p'_0(\vx) } \nonumber \\ & + \E[\vx \sim p]{ \log \E[\vx' \sim p'_0]{\exp \g(\vx)} - \log \E[\vx' \sim p''_0]{\exp \g(\vx)}  } \nonumber \\
&= \mathrm{KL}[p||p'_0]-\mathrm{KL}[p||p''_0] \nonumber \\ & + \log \frac{\E[\vx' \sim p'_0]{\exp \g(\vx)}}{\E[\vx' \sim p''_0]{\exp \g(\vx)}}. \label{eq:mild}
\end{align}
Plugging assumptions $\frac{\E[\vx \sim p'_0]{\exp \g(\vx)}}{ \E[\vx \sim p''_0]{\exp \g(\vx) }  } > \exp(-k)$ and ${\mathrm{KL}} [p || p'_{0}] - {\mathrm{KL}} [p || p''_{0}] > k$ into \cref{eq:mild}, $\mathrm{KL}[p || p'_{\vtheta}] - \mathrm{KL}[p || p''_{\vtheta}] > 0$.
\end{proof}

\Cref{propo:bound} suggests a training strategy that we first train the base model $p_0$, then finetune $\g$: under a roughly uniform $\g$ (\emph{e.g.} when $\vtheta$ is newly initialized), $\nicefrac{\E[\vx \sim p'_0]{\exp \g}}{\E[\vx \sim p''_0]{\exp \g}} \approx \exp(0)$; so
 improvements on the inclusive $\mathrm{KL}$-divergence of base model $\mathrm{KL}[p||p_0]$ will mostly translate to improvement in $\mathrm{KL}[p||\tilde{p}_{\vtheta}]$. 
 Optimizing the base model (\emph{i.e.} finding $p''_0$ such that $\mathrm{KL}[p||p''_0] < \mathrm{KL}[p||p''_0]$) is much easier than directly minimizing $\mathrm{KL}[p||p'_{\vtheta}]$: the former can be done by minimizing empirical cross entropy, which is computationally efficient, while the latter involves an intractable partition function $\sum_{\vx \in \Xc_{\leq T}} \tilde{p}'_{\vtheta}(\vx)$.
 
 Pseudocode for fine-tuning $\g$ is listed in \cref{alg:training-g}. 

\todo{try connect $g_{optimal}(\vx) = \log p^*(\vx) - \log p_0(\vx) + C$ and the optimal classifier}
\begin{algorithm}[h]
\SetAlgoLined
\KwIn{
\begin{itemize}
\item Training/validation corpora $\mathcal{D}_{\{\text{train}, \text{dev}\}}$ 
\item  base model $p_0: \Xc_{\leq T} \rightarrow [0, 1]$ 
\item initial parameter vector $\vtheta_0 \in \Bool^d$ \item noise sample size $K \in \mathbb{N}$
\end{itemize}}
\KwOut{unnormalized residual language model $\tilde{q}_{\vtheta}: \Xc_{\leq T} \rightarrow [0, 1]$}
$\vtheta \leftarrow \vtheta_0$ \;
 \tcc{$\mathcal{L}_{\textsc{nce}}$ is defined in \cref{eq:rlm-nce-loss}}
 \While{$\sum_{\vx \in \mathcal{D}_{\text{dev}}} \mathcal{L}_{\textsc{nce}}(\vtheta, \vx, p_0, K)$ is still decreasing}{
 \ForEach{$\vx \in \text{shuffle}(\mathcal{D}_{\text{train}})$}{
 $\nabla_{\vtheta} \mathcal{L}_{\textsc{nce}} = \nabla_{\vtheta} \mathcal{L}_{\textsc{nce}} (\vtheta, \vx, p_0, K )$\;
 $\vtheta \leftarrow \text{update-gradient}(\vtheta, \nabla_{\vtheta} \mathcal{L}_{\textsc{nce}})$\;
 }
 }
 \Return{$ \vx \mapsto p_0(\vx) + \exp \g(\vx) $}\;
 \caption{Pseudocode for training $\g$}
 \label{alg:training-g}
\end{algorithm}
\subsection{Computing normalized probabilities}
\label{sec:computing-normalized}
The unnormalized probability $\ptilde_{\vtheta}(\vx)$ (in \cref{eq:residual-sum}) can be evaluated easily, and should suffice for (re)ranking purposes (\emph{e.g.} for ASR and MT applications). However, the \emph{normalized} probability $q_{\vtheta}(\vx) \triangleq \frac{\ptilde_{\vtheta}(\vx)}{ \sum_{\vx} \ptilde_{\vtheta}(\vx)  }$ does require computing the partition function $\Z$. An unbiased importance sampling estimate of $\sum_{\vx \in \Xc_{\leq T}} \ptilde_{\vtheta}(\vx)$ is
\begin{align}
    \Z &= \sum_{\vx \in \Xc_{\leq T}} \ptilde_{\vtheta}(\vx) \nonumber \\
    &= \sum_{\vx \in \Xc_{\leq T}} p_0(\vx) \exp \g(\vx) \nonumber \\
    &= \E[\vx \sim p_0]{ \exp \g(\vx) } \nonumber \\
    & \approx \sum_{m=1}^{M} \frac{ \exp \g(\vx^{(m)}) }{M} = \hat{\Z}_M,
    \label{eq:z-estimate}
\end{align}
where $\vx^{(1)} \ldots \vx^{(M)} \sim q_{0}$.
\section{Comparison between REBMs and autoregressive models}
\label{sec:comparison-local}
\begin{table*}[h]
	\scriptsize
	\centering
	\begin{tabular}{p{.2\linewidth}p{.06\linewidth}p{.15\linewidth}rr}
		\toprule
		Experiment (Architecture) & Model & Best configuration & log likelihood improvement ($95\%$ CI) & perplexity improvement \\ \midrule
		RealNews (Transformer) & $p_{\vtheta}$ & $4$-layer, $\tanh$ & $(\num{-0.17727623084440403}, \num{-0.13434780338659458}), \mu = \num{-0.15492809475605884}$ &  $.03\%$  \\
		RealNews (Transformer) & $p_0'$  & N/A  & N/A %
		& $.00\%$\\
		\midrule
		WikiText (GRU) & $p_{\vtheta}$ & $1$-layer/$500$, softplus & $(\num{-1.849087617317963}, \num{-1.535010239998627}), \mu = \num{-1.69034763930254}$ &  $1.44\%$  \\
		WikiText (GRU) & $p_0'$ & N/A & N/A &  $.50\%$  \\
		\midrule
		Yelp (GRU) & $p_{\vtheta}$ &  $2$-layer/$500$, softplus & $(\num{-1.890286347786713}, \num{-1.6666764234703066}), \mu = \num{-1.7954674374859811}$ &  $1.82\%$  \\
		Yelp (GRU) & $p_0'$ & N/A & N/A &  $.49\%$  \\
		\bottomrule
	\end{tabular}
	\caption{\small Residual energy-based model $\tilde{p}_{\vtheta}$ improvements over autoregressive base models $p_0$. The perplexity numbers are per-token, and log likelihood improvements are per sequence (in nats).  We only report each dataset's best model (according to validation data) in this table. See \cref{sec:experimental-details} for experimental details.}
	\label{tbl:residual-improvements-brief}
\end{table*}
We evaluate the effectiveness of REBMs on two different neural architectures (GRU- and Transformer-based) and 3 datasets: WikiText \cite{Merity2017PointerSM}, Yelp \cite{yelp}, and RealNews \cite{zellers2019defending}, on the task of modeling sequence probabilities.  An REBM $\tilde{p}_{\vtheta}$ has two components, $\g$ and $p_0$, and we would like to see how $\tilde{p}_{\vtheta}$ competes against $p_0$ itself.  We do not further tune $p_0$ while training $p_{\vtheta}$.  As a fair comparison, we also see how $p_0'$ compares against $p_0$, where $p_0'$ is simply a version of $p_0$ that has been trained as many additional epochs as were used to train $p_{\vtheta}$. 

$p_0$ models are pretrained on moderately large corpora (in GRU cases) or a very large corpus (in the Transformer case).\footnote{In the Transformer case we simply take $p_0$ to be the Grover \cite{zellers2019defending} pretrained language model, which is based on the GPT-2 \cite{radford2019language} architecture and performs competitively on news article generation.}
We compare residual energy-based models $\tilde{p}_{\vtheta}$ to further-fine-tuned base models $p_0'$, on conservatively estimated (at the low end of $95\%$ confidence interval) token perplexity and bootstrap-sampled log likelihood improvements. The results are in \cref{tbl:residual-improvements-brief}.
Residual energy-based models show consistent perplexity improvement compared to  $p_0'$ that are trained on the same data using the same maximum numbers of iterations.
Although the improvement in log-likelihood of $p_{\vtheta}$ over $p_0$ is modest (especially for RealNews experiments, where $p_0$ is a very strong baseline), we verify that these improvements are all statistically significant ($p < 0.05$) using bootstrapped test datasets.

We experiment with different designs of the discriminator $\g$, evaluating the effectiveness of bounding $\g$ and varying its number of parameters. We find that in Transformer-based experiments, bounding $\g$ considerably helps with performance; but the opposite happens for GRU-based models. We speculate that this is due to the base models' performance: the Transformer base models have high parameter count and were trained on a lot of data; and the true distribution $p$ likely is relatively similar to $p_0$, and benefits from a small hypothesis space \Dash even though we don't know if the at-most-$\epsilon$ error assumption in \cref{sec:ebm} holds. On the other hand our GRU-based $p_0$ has neither the capacity, nor the huge amount of training data. As a result, the unbounded variant $\g$ (and ${q}_{\vtheta}$) may end up learning a better approximation of $p$.

\section{Experimental details}
\label{sec:experimental-details}
\subsection{Datasets}
\label{sec:datasets}
Residual language model experiments are conducted on these datasets:
\begin{itemize}
    \item {\bf Segmented WikiText}: we take the standard WikiText-2 corpus \cite{Merity2017PointerSM}, and segment it into sequences at new line breaks. We discard all empty lines, and any line that starts with the `\texttt{=}' token. In effect, we obtain sequences that are mostly entire paragraphs. We also only keep lines that are shorter than $800$ tokens after BPE tokenization. Because of our preprocessing, Segmented WikiText loses much interparagraph context information, and doesn't have the `simple' header sequences that were in the original WikiText corpus, and is much harder to language-model.
    \item{\bf Yelp}: the Yelp dataset \cite{yelp} contains business reviews. As in Segmented WikiText, We keep reviews shorter than $800$ tokens.
    \item {\bf \textsc{RealNews}}: we make use of the standard \textsc{RealNews} corpus comes from \cite{zellers2019defending}, which contains news articles that are up to $1,024$ tokens long.
\end{itemize}
In all experiments we tokenize with BPE tokenizers derived from the GPT-2 language models: the GRU models use Huggingface's implementation\footnote{\url{https://github.com/huggingface/transformers}} and the Transformers use Grover's\footnote{\url{https://github.com/rowanz/grover}}.
Number of sequences in preprocessed datasets are listed in \cref{tbl:dataset-sizes}.
\begin{table}[h]
\centering
\begin{tabular}{lrrr}
\toprule
         & Train & Dev & Test \\ \midrule
RealNews &   $3,855$    &  $1,533$   & $6,158$     \\
WikiText & $18,519$      &  $878$   & $2,183$      \\ 
Yelp     &   $10,951$    &  $9,964$    &    $994$  \\ \bottomrule
\end{tabular}
\caption{Number of sequences in preprocessed datasets (for training and tuning the discriminators $\g$, and evaluation).}
\label{tbl:dataset-sizes}
\end{table}
\subsection{Pretraining base models $p_0$}
We use a pretrained Grover model as the base model in RealNews experiments. For GRU-based experiments, we train base models on  WikiText and Yelp datasets using separate training and validation splits than those of the discriminator $\g$ (\cref{tbl:base-dataset-sizes}). The base models are periodically (every $1,000$ iterations) evaluated on the validation split for early stopping, where we stop if there is no improvement on validation perplexity for $10$ consecutive evaluations. The base models $q_{\vtheta}$ achieve $\num{113.9793886764}$ for Segmented WikiText, and $\num{110.8903443378}$ in test set perplexity, respectively. Note that these base models are further fine-tuned on additional datasets in our comparison against residual language models.
\begin{table}[h]
\centering
\begin{tabular}{lrr}
\toprule
         & Train & Dev  \\ \midrule
WikiText & $17,556$      &  $1,841$    \\ 
Yelp     &   $9,954$    &  $1,000$   \\ \bottomrule
\end{tabular}
\caption{Number of sequences in preprocessed datasets (for training and tuning the base model $q$). Note that we do not train our own base models for RealNews, but use one of the pretrained models provided by \cite{zellers2019defending}.}
\label{tbl:base-dataset-sizes}
\end{table}
\subsection{Metrics}
We evaluate the relative performance of residual language models against autoregressive models (\emph{i.e.} fine-tuned base models) on two metrics, log likelihood and perplexity improvement, which are approximated as follows:
\begin{itemize}
    \item {\bf Log likelihood improvement}: since $p$, $p_{\vtheta}$ and $q_{0}$ are all distributions over $\Xc_{\leq T}$, we can quantitatively evaluate their difference in log likelihood. We measure the difference between $\mathrm{KL}[p||p_{\vtheta}]$ and $\mathrm{KL}[p||p_0]$:\footnote{Note that $p_0$ here is the base model component of $\tilde{p}_{\vtheta}$. While comparing between residual language models and autoregressive models, we also finetune $p_0$  on additional data to get a new model $q'_0$, which has different parameters than $p_0$.}
    \begin{align}
        & \mathrm{KL}[p||{p}_{\vtheta}] - \mathrm{KL}[p||p_0] \nonumber \\ &= \E[\vx \sim p]{ \log {p}_{\vtheta}(\vx) - \log p_0(\vx) } \nonumber \\
        &= \E[\vx \sim p]{\log \tilde{p}_{\vtheta}(\vx) - \log p_0(\vx)} \nonumber - \log \Z \nonumber \\
        &= \E[\vx \sim p]{ \g(\vx)} - \log \Z \nonumber \\
        &\approx \frac{\sum_{\vx \in \mathcal{D}_{\text{test} }} \g(\vx)}{|\mathcal{D}_{\text{test}}|} - \log \hat{\Z}_M,
        \label{eq:ll-improvement}
    \end{align}
    where $\hat{\Z}_M$ is estimated using \cref{eq:z-estimate}. A negative value of log likelihood difference indicates that $\tilde{q}_{\vtheta}$ approximates $p$ better than $p_0$ in terms of $\mathrm{KL}$-divergence.
    \item {\bf Perplexity improvement}: perplexity is a common language modeling metric. Following \cite{rosenfeld2001}, we compute
    \begin{align}
        &\text{perplexity improvement of } p_{\vtheta} \nonumber \\ &= \frac{\exp  \frac{|\mathcal{D}| \log \hat{\Z}_M}{ w(\mathcal{D}_{\text{test}}) }}{ \exp \frac{ \sum_{\vx \in \mathcal{D}_{\text{test}}}  \g(\vx) }{ w(\mathcal{D}_{\text{test}}) }},
        \label{eq:ppl-improvement}
    \end{align} where $w(\mathcal{D})$ is the total token count of dataset $\mathcal{D}$, and $|\mathcal{D}|$ is the number of sequences of $\mathcal{D}$. $\hat{\Z}_M$ is ecomputed \cref{sec:computing-normalized}
\end{itemize}
Both evaluation metrics involve estimating the partition function with $\hat{\Z}_{M}$. For the perplexity improvement metric, we obtain $32$ estimates of $\hat{\Z}_M$\footnote{We set $M=512$ in this paper.}, which are normally distributed, and compute \cref{eq:ppl-improvement} using $\hat{\Z}_M$ the conservative end of a $95\%$ confidence level.
To account for variance in our test datasets, we further make use of bootstrapping estimation for log likelihood improvement: we bootstrap-sample $1,000$ subsamples for each test dataset, and compute \cref{eq:ll-improvement} for each datapoint in the Cartesian product ($1,000 \times 32$ in total). We then report results at the $2.5\%$ and $97.5\%$ percentiles.
\subsection{Hyperparameters}
\paragraph{Transformer experiments.} We train our models on $64$ GPUs across $8$ nodes, with a total batch size of $64 \times 8 \times 2 = 1,024$, and with $1$ noise sequence ($K=1$ in \cref{sec:training-procedure}) per batch. We use an initial learning rate of $5e-5$. The rest of the hyperparameters largely follow settings in \cite{zellers2019defending}. Optimization is done with the Grover implementation of AdaFactor.
\paragraph{GRU experiments.} We train our models on $8$ GPUs on a single node, with a total batch size of $8 \times 2 = 16$, and with $25$ noise sequences ($K=25$ in \cref{sec:training-procedure}) per batch. We have an initial learning rate of $1e-4$. Upon no improvement on validation data, we half the learning rate, with patience $=1$. The model parameters are $l_2$ regularized with a coefficient of $1e-5$. We also apply dropout regularization with $p=0.5$. Optimization is done with PyTorch-supplied Adam.
\subsection{Configurations}
We study the effects of these configurations:
\begin{itemize}
    \item {\bf Bounding $\g$}: we note in \cref{sec:ebm} that with the strong  hypothesis that the base model $p_0$ has bounded error, $\g$ will have a bounded range, and leads to a much smaller hypothesis space. In this work we experiment with both bounded and unbounded $\g$'s, with ranges $(-\infty, 0)$ and $(-2, 2)$ respectively. More details can be found in \cref{sec:discriminator-design}.
    \item {\bf Model capability of $\g$}: we hypothesize that the expressiveness of $\g$ does not need to be as rich as the parametrization of $p_0$, since $\g$ essentially only has to tell whether the sequence $\vx$ comes from $p$ or $p_0$. For the GRU + WikiText experiments, we experiment with $\{1, 2\}$-layer GRU models of $\g$. For $1$-layer models, we additionally experiment with a setup that has only $250$ hidden units. For the Transformers/RealNews dataset, we experiment with $\{12, 4\}$-layer Transformer models.
\end{itemize}
\subsection{Log likelihood improvements under different configurations}
\begin{table*}[h]
\centering
\begin{tabular}{ccrr}
\toprule
\multirow{2}{*}{Model Size} & \multirow{2}{*}{Activation} & \multicolumn{2}{c}{log likelihood improvement} \\ \cmidrule{3-3} \cmidrule{4-4} 
 &  & \multicolumn{1}{c}{$95\%$ CI}  & \multicolumn{1}{c}{$\mu$}              \\ \midrule
\multicolumn{4}{c}{RealNews (Transformers)}                \\ \midrule
$12$-layer  & softplus    & $(\num{-0.13125226596250883}, \num{0.07534410854920992})$ & $\num{-0.09380679199352614}$ \\
$12$-layer  & tanh       & $(\num{-0.1416431138838785}, \num{-0.09855092266454868})$ & $\num{-0.11840874034046701}$ \\
$4$-layer  & softplus      & $(\num{-0.1486568222369229}, \num{2.617413543668839})$ & $\num{-0.023575736435798304}$ \\
$4$-layer  & tanh    & $(\num{-0.17727623084440403}, \num{-0.13434780338659458})$ & $\num{-0.1550275131875793}$ \\
 \midrule
\multicolumn{4}{c}{WikiText (GRUs)}                        \\ \midrule
$2$-layer / 500  & tanh     & $(\num{-0.0035970424812301616}, \num{0.002045013983918764})$ & $\num{-0.0009496680002674082}$ \\
$2$-layer / 500 & softplus       & $(\num{-1.3230198835533145}, \num{-0.845958611885834})$   & $\num{-1.1750736540358546}$   \\
$1$-layer / 500 & tanh     &  $(\num{-0.7913531278770431}, \num{-0.6401689266365036})$ & $\num{-0.7122775004960611}$ \\
$1$-layer / 500 & softplus       & $(\num{-1.849087617317963}, \num{-1.535010239998627})$   & $\num{-1.69011571661644}$  \\
$1$-layer / 250 & tanh       &  $(\num{-0.022264740387724302}, \num{0.015675523360444643})$  & $\num{-0.0036657970722779256}$ \\
$1$-layer / 250 & softplus  &  $(\num{-1.8494223570030215}, \num{-1.4615611051719668})$ & $\num{-1.6713931474488262}$ \\
\midrule
\multicolumn{4}{c}{Yelp (GRUs)}                        \\ \midrule
$2$-layer / 500  & tanh     & $(\num{-0.02805521241836395}, \num{0.010745265563203432})$ & $\num{-0.016728270265088937}$   \\
$2$-layer / 500 & softplus       & $(\num{-1.890286347786713}, \num{-1.6666764234703066})$    & $\num{-1.7953834329050067}$  \\
$1$-layer / 500 & tanh     &  $(\num{-0.649808666626738}, \num{-0.5714733814399704})$ & $\num{-0.6140814270328745}$ \\
$1$-layer / 500 & softplus       & $(\num{-2.6185950254600527}, \num{-2.028866670052338})$ & $\num{-2.4270503117244724}$    \\
$1$-layer / 250 & tanh       &  $(\num{-0.004170081536100767}, \num{0.0007728124458328267})$  & $\num{-0.0014251322213574866}$ \\
$1$-layer / 250 & softplus  &  $(\num{-2.248356721321869}, \num{-1.9872111295860293})$ & $\num{-2.127174716333676}$ \\
\bottomrule
\end{tabular}
\caption{Comparison of different configurations.}
\label{tbl:config-comparison}
\end{table*}

We also see in \cref{tbl:config-comparison} that using $\tanh$ as the activation function $f$ does better than softplus for Transformers; but performs very poorly for GRUs. We also observe degeneracy problems. We speculate that our Transformer-based base models $q_{\vtheta}$ have already learned a good approximation of the true distribution; and limiting the model capacity of $\g$ in exchange of smaller variance results in a favorable trade-off, and vice versa for GRUs. 
Regarding discriminator capability: we see that performance is not sensitive to model size. Our best Transformers run actually is from the smaller-model runs. And the 1-layer 500-unit GRU models achieve best performance. Overall, results in \cref{tbl:config-comparison} suggests that performance is sensitive to the choice of model configuration.